\documentclass[10pt,journal,compsoc]{IEEEtran}
\usepackage{amsmath,amsfonts}
\usepackage{algorithmic}
\usepackage{algorithm}
\usepackage{array}
\usepackage{textcomp}
\usepackage{stfloats}
\usepackage{url}
\usepackage{verbatim}
\usepackage{graphicx}
\usepackage{todonotes}
\usepackage{hyperref}
\usepackage[capitalise]{cleveref}
\usepackage{multirow}
\usepackage{array}
\usepackage{eccvabbrv}
\usepackage{booktabs}
\usepackage{subcaption}
\usepackage[style=ieee,maxbibnames=5,mincitenames=1,maxcitenames=1]{biblatex}

\definecolor{figblue}{RGB}{40,96,144}
\definecolor{figorange}{RGB}{216,109,34}
\definecolor{figgreen}{RGB}{48,130,44}
\definecolor{figctp}{RGB}{39,96,144}
\definecolor{figlunitdino}{RGB}{52,130,45}

\usepackage[acronym,shortcuts]{glossaries}
\glsdisablehyper
\newacronym{wsi}{WSI}{whole slide image}
\newacronym{ssl}{SSL}{self-supervised learning}
\newacronym{auroc}{AUROC}{area under the receiver operating characteristic curve}
\newacronym{he}{H\&E}{haematoxylin and eosin}
\newacronym{cnn}{CNN}{convolutional neural network}
\newacronym{mil}{MIL}{multiple instance learning}
\newacronym{tcga}{TCGA}{The Cancer Genome Atlas}
\newacronym{srcl}{SRCL}{semantically-relevant contrastive learning}
\newacronym{ccl}{CCL}{clustering-guided contrastive learning}
\newacronym{ood}{OOD}{out of distribution}
\newacronym{crc}{CRC}{colorectal cancer}
\newacronym{msi}{MSI}{microsatellite instability}
\newacronym{cpath}{CPATH}{computational pathology}
\newacronym{mpp}{MPP}{microns per pixel}

\usepackage{scalerel,graphicx,xparse}
\NewDocumentCommand\colonicon{}{\scalerel*{\includegraphics{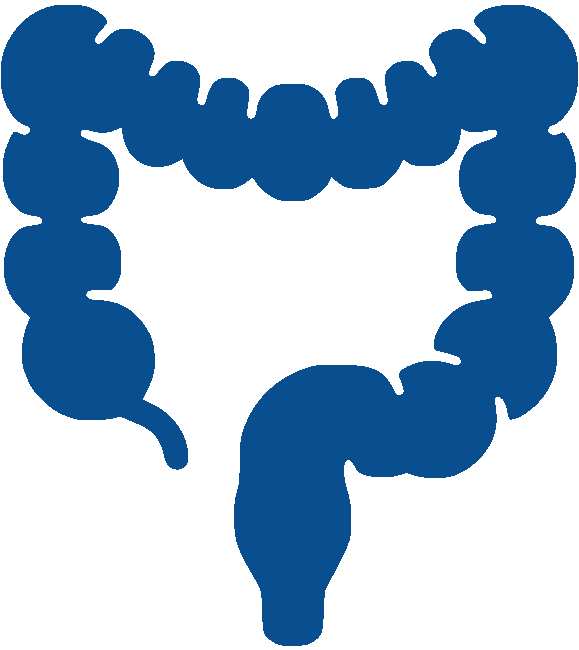}}{X}}
\NewDocumentCommand\breasticon{}{\scalerel*{\includegraphics{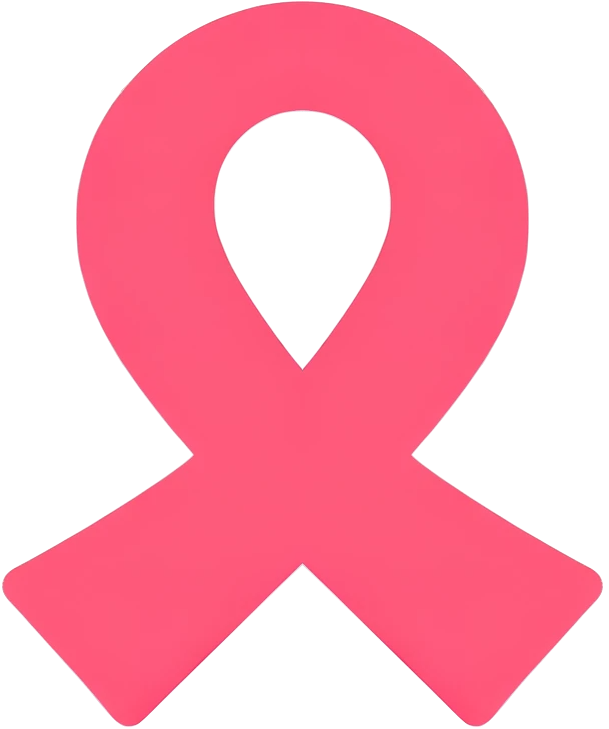}}{X}}

\begin{document}

\title{Benchmarking Pathology Feature Extractors for Whole Slide Image Classification}

\author{%
  Georg~W\"{o}lflein, 
  Dyke~Ferber, 
  Asier~R.~Meneghetti, 
  Omar~S.~M.~El~Nahhas, 
  Daniel~Truhn,\\
  Zunamys~I.~Carrero,
  David~J.~Harrison,
  Ognjen~Arandjelovi\'{c},
  Jakob~Nikolas~Kather%
\IEEEcompsocitemizethanks{%
\IEEEcompsocthanksitem G.W., D.J.H., O.A.\ are with the University of St Andrews, United Kingdom. 
\IEEEcompsocthanksitem G.W., D.F., A.R.M., O.S.M.E.N., Z.I.C., J.N.K.\ are with Else Kr\"{o}ner Fresenius Center for Digital Health, Medical Faculty Carl Gustav Carus, TUD Dresden University of Technology, Dresden, Germany. 
\IEEEcompsocthanksitem D.T.\ is with University Hospital Aachen, Germany. 
\IEEEcompsocthanksitem D.F.\ and J.N.K.\ are with Department of Medical Oncology, National Center for Tumor Diseases (NCT), University Hospital Heidelberg, Heidelberg, Germany.
\IEEEcompsocthanksitem J.N.K.\ is with Department of Medicine I, University Hospital Dresden, Dresden, Germany.}
\thanks{Code and data are available at \protect\url{https://georg.woelflein.eu/good-features}.}%
\thanks{Manuscript received June 20, 2024.}}

\markboth{Preprint, under review.}%
{W\"{o}lflein \MakeLowercase{\textit{et al.}}: Benchmarking Pathology Feature Extractors for Whole Slide Image Classification}


\IEEEtitleabstractindextext{%
\begin{abstract}
  Weakly supervised whole slide image classification is a key task in computational pathology, which involves predicting a slide-level label from a set of image patches constituting the slide.
  Constructing models to solve this task involves multiple design choices, often made without robust empirical or conclusive theoretical justification.
  To address this, we conduct a comprehensive benchmarking of feature extractors to answer three critical questions:
  1) Is stain normalisation still a necessary preprocessing step?
  2) Which feature extractors are best for downstream slide-level classification?
  3) How does magnification affect downstream performance?
  Our study constitutes the most comprehensive evaluation of publicly available pathology feature extractors to date, involving more than 10,000 training runs across 14 feature extractors, 9 tasks, 5 datasets, 3 downstream architectures, 2 levels of magnification, and various preprocessing setups.
  Our findings challenge existing assumptions: 
  1) We observe empirically, and by analysing the latent space, that skipping stain normalisation and image augmentations does not degrade performance, while significantly reducing memory and computational demands. 
  2) We develop a novel evaluation metric to compare relative downstream performance, and show that the choice of feature extractor is the most consequential factor for downstream performance.
  3) We find that lower-magnification slides are sufficient for accurate slide-level classification.
  Contrary to previous patch-level benchmarking studies, our approach emphasises clinical relevance by focusing on slide-level biomarker prediction tasks in a weakly supervised setting with external validation cohorts. 
  Our findings stand to streamline digital pathology workflows by minimising preprocessing needs and informing the selection of feature extractors.
\end{abstract}

}

\maketitle

\IEEEdisplaynontitleabstractindextext

%
\IEEEpeerreviewmaketitle

\begin{figure*}[t]
  \centering
  \includegraphics[width=.8\linewidth]{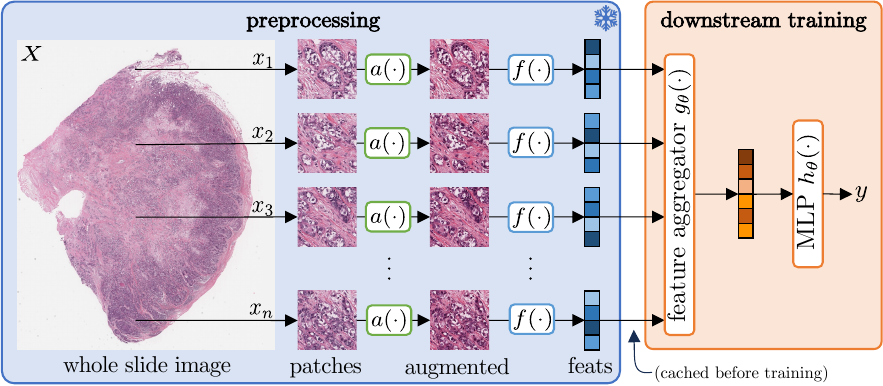}
  \caption{Common setup for weakly supervised learning on \glspl{wsi}. In the preprocessing stage, the input image is split into patches that undergo independent image augmentations $a$ before feature extraction. The feature aggregator and classifier are trained jointly as a single neural network $g_\theta \circ h_\theta$, which, given feature vectors as inputs, predicts the output $y$. For stain normalisation (shown here), the same $a(\cdot)$ is applied every time, though in general, the augmentation function may vary between patches and epochs.}
  \label{fig:overview}
\end{figure*}

\IEEEraisesectionheading{\section{Introduction}\label{sec:introduction}}
\IEEEPARstart{T}{here} has been a surge in studies using deep learning in oncology to predict clinical variables such as genetic alterations and survival directly from routinely available histopathology \glspl{wsi}~\cite{coudray2018classification,kather2019deep,lu2021dataefficient,wagner2023transformerbased,ghaffari2022benchmarking,niehues2023generalizable,elnahhas2024regressionbased,yang2023devil,loeffler2023direct,elnahhas2023whole,liu2023identification}. 
Due to their immense size reaching billions of pixels, these images are first divided into small, non-overlapping patches, followed by a two-step process involving 
  (i) feature extraction, where a feature vector is obtained separately for each patch and
  (ii) feature aggregation, where the extracted feature vectors are combined to form the slide-level prediction~\cite{campanella2019clinical,shmatko2022artificial},
as illustrated in \cref{fig:overview}.
Both steps are parametrised using neural networks; usually, the feature extractor is a deep backbone architecture whose parameters are frozen, while the aggregator is shallower, but trainable. 
Keeping the feature extractor frozen allows all feature vectors to be pre-computed before training, so that the downstream training process is computationally feasible.
In the past, \glspl{cnn} such as ResNet-50~\cite{he2015deep} pretrained on ImageNet~\cite{deng2009imagenet} were used for feature extraction. 

Recent advances in \gls{ssl} make it possible to train powerful feature extractors without labels, a development that is gaining traction in computational pathology, where large quantities of images are available but annotations are sparse. 
The last few years have witnessed the emergence of several \gls{ssl} models trained on large-scale pathology datasets~\cite{wang2021transpath,chen2022scaling,wang2022transformer,chen2024uni,azizi2023robust,wang2023retccl,filiot2023scaling,kang2023benchmarking,vorontsov2023virchow,lu2024conch}. 
These models produce better representations for downstream tasks than their ImageNet-pretrained counterparts~\cite{dehaene2020selfsupervision,ciga2022self,kang2023benchmarking,campanella2023computational,chen2021selfsupervisedhistopathology,sikaroudi2023generalization}, and are establishing themselves as the leading choice for feature extraction~\cite{guan2022node,wagner2023transformerbased,elnahhas2024regressionbased,niehues2023generalizable,wolflein2023deep,saldanha2023selfsupervised,xiang2023automatic}.

For over two decades, stain normalisation~\cite{reinhard2001color,macenko2009method} has been a standard preprocessing step in computational pathology pipelines to account for variations in scanners and \gls{he} stains by adjusting \glspl{wsi} to match a reference image. 
It remains an active area of research~\cite{wagner2021structure,nazki2023multipathgan,zanjani2018stain,vahadane2016structure}, and is widely adopted~\cite{ghaffari2022benchmarking,chikontwe2022weakly,schrammen2022weakly,elnahhas2024regressionbased,elnahhas2023whole,liu2023identification} in weakly supervised \gls{wsi} classification.
Yet, with the shift from ImageNet \glspl{cnn} to \gls{ssl} models trained on vast and varied pathology data from multiple centres, it is worth reconsidering its need. 
Beyond stain normalisation, image augmentations are a broad category of image-to-image transformations that may be applied during training, such as random flips, rotations, and colour transformations. 
Some augmentations, like rotation, are particularly well-suited for pathology due to the rotational invariance of micrographs~\cite{salvi2021impact}. 
\Gls{ssl} feature extractors that have been trained on a wide variety of images from multiple international sites might therefore extract diagnostically/prognostically relevant features irrespective of site- or scanner-specific traits. 
This leads to our primary research question: \emph{with \gls{ssl} feature extractors trained on rich datasets, is there still a need for image augmentations and stain normalisation to improve the generalisability of weakly supervised whole slide image classification models?}
Our study approaches this question in two ways:
\begin{enumerate}
    \item We assess the latent space similarity between original patches and their stain-normalised/augmented counterparts in \cref{sec:latent}. 
          Our analysis reveals that many augmentations induce only minor perturbations in the extracted features, especially compared to ImageNet backbones.
    \item In the most comprehensive robustness evaluation of publicly available pathology \gls{ssl} feature extractors to date, we compare over 10,000 trained models, with and without normalisation/augmentation, across multiple externally validated slide-level tasks.
          We find (i) the choice of feature extractor is most consequential for downstream performance, (ii) stain normalisation and image augmentations have minimal effect, and (iii) the leading feature extractors demonstrate comparable effectiveness at both low ($\approx 9\times$) and high ($20\times$) magnification levels.
\end{enumerate}

The goal of this paper is \emph{not} to propose a new architecture. 
Instead, we aim to challenge the long-standing belief that stain normalisation is necessary for weakly supervised \gls{wsi} classification, and to identify the best publicly available feature extractors.
Our findings have implications for computational pathology researchers and practitioners alike, given that stain normalisation is an active research area~\cite{wagner2021structure,nazki2023multipathgan,zanjani2018stain,vahadane2016structure} and that it incurs substantial computational overhead in pathology pipelines.
We stress that our findings are specific to weakly supervised \gls{wsi} classification, which is arguably one of the most important tasks within computational pathology, and we make no claims about other tasks.

Parts of this paper were previously published as a conference paper~\cite{wolflein2023good}.
In this extended version, we
(i) broaden our analysis to include experiments at a higher magnification level (0.5 \gls{mpp} in addition to 1.14 \gls{mpp} in the original paper), and 
(ii) include two additional feature extractors, UNI~\cite{chen2024uni} as well as an ImageNet-pretrained~\cite{deng2009imagenet} baseline of its underlying ViT-L~\cite{kolesnikov2021image} architecture. 
UNI~\cite{chen2024uni} constitutes an interesting addition because it was trained on an order of magnitude more slides than previosuly available pathology feature extractors, and is the first pathology foundation model to release its weights to the research community.

\subsection{Problem formulation}
In a slide classification task, we have a dataset of labelled \glspl{wsi}. 
Each \gls{wsi} $X\in\mathbb{R}^{W \times H \times 3}$ is an RGB image whose dimensions $W$ and $H$ may vary between slides. 
It is associated with a ground truth label $y \in \mathcal{Y} = \mathbb{R}^c$ for a $c$-way classification problem. 
Due to their large size, we consider each \gls{wsi} as a bag of patches, framing the \gls{wsi} classification problem as a weakly supervised learning task. 
More specifically, we split each \gls{wsi} $X$ into a set of $n$ non-overlapping patches $\{x_1,x_2,\dots,x_n\}$ where each $x_i\in\mathcal{X}=\mathbb{R}^{P\times P\times 3}$ for fixed patch size $P$. 
Here, $n$ varies depending on the particular slide's dimensions (usually between 1,000 and 10,000 at $10\times$ magnification with $P=224$).
The task is to find a model $M : \mathcal{X}^n\to\mathcal{Y}$ that predicts the label given a bag of patches representing a \gls{wsi}.

It is computationally infeasible to parametrise $M$ using a single deep neural network trained end-to-end. 
Instead, the common approach in the literature is a two-step process consisting of preprocessing (feature extraction) and training (aggregation and classification), outlined in \cref{fig:overview}.
The preprocessing stage often entails stain normalisation, and sometimes includes image augmentations as well.

We first consider the simple case with a predetermined augmentation function $a : \mathcal{X}\to \mathcal{X}$ that is applied independently to each patch $x_i$ to obtain the augmented patches $\hat{x}_i = a(x_i)$ for $i=1,2,\dots,n$. 
Then, we apply the feature extractor $f: \mathcal{X} \to \mathbb{R}^{d_x}$, which for each patch $\hat{x}_i$ outputs a $d_x$-dimensional feature vector $z_i=f(\hat{x}_i)$. 
Now, we have $n$ feature vectors, $z_1,z_2,\dots,z_n$, which are aggregated into a single vector $\bar{z}\in\mathbb{R}^{d_z}$ (usually $d_x=d_z$) via an aggregation function $g_\theta : \mathbb{R}^{n\times d_x} \to\mathbb{R}^{d_z}$ with learnable parameters $\theta$. 
Finally, the aggregated feature vector $\bar{z}$ passes through a classifier $h_\theta: \mathbb{R}^{d_z} \to \mathcal{Y}$, to obtain the final prediction. 
In summary, we can express the process $M: \mathcal{X}^n\to \mathcal{Y}$ of obtaining a prediction $y$ from a bag of patches $\{x_i\}_{i=1}^n$ as
\setlength{\abovedisplayskip}{2pt}
\setlength{\belowdisplayskip}{2pt}
\begin{equation}
    \label{eq:mil}
    M(\{x_i\}_{i=1}^n) = \underbrace{(h_\theta \circ g_\theta)\overbrace{\left(\{(f \circ a)(x_i)\}_{i=1}^n\right)}^\mathrm{preprocessing}}_\mathrm{training},
\end{equation}
where $\circ$ denotes function composition. Notice that $f\circ a$ is independent of the learnable parameters $\theta$; it can be pre-computed for all patches $x_i$ before training. 

In the general case, we define a set of augmentation functions $\mathcal{A} \in \mathcal{X}^\mathcal{X}$ before training%
\footnote{%
    This formulation permits any augmentation we may wish to perform.
    The set of augmentations $\mathcal{A}$ may contain, for example, a function that rotates the input image. 
    If, during training, we want to rotate each patch by a random angle, we can simply define a rotation function for every angle, and include all of these functions in $\mathcal{A}$. 
}
($\mathcal{X}^\mathcal{X}$ is the set of functions from $\mathcal{X}$ to $\mathcal{X}$). During training, for every patch $x_i$, we uniformly sample\footnote{The augmentation is resampled for every patch at every epoch.} an augmentation $a_i \sim \mathcal{A}$. 
Then, the augmented feature vector is $\hat{x}_i = a_i(x_i)$, so \cref{eq:mil} becomes
\setlength{\abovedisplayskip}{4pt}
\setlength{\belowdisplayskip}{4pt}
\begin{equation}
    \label{eq:mil_aug}
    M(\{x_i\}_{i=1}^n) = (h_\theta \circ g_\theta)\left(\{(f \circ a_i)(x_i)\}_{i=1}^n\right).
\end{equation}
While just a small modification in terms of notation, this change incurs a significant increase in time and memory complexity of the preprocessing task by a factor of $|\mathcal{A}|$, since augmentation and feature extraction must be performed for all possible augmentations $a_i \in\mathcal{A}$ for every patch\footnote{If there are fewer augmentations $|\mathcal{A}|$ than epochs, it is cheaper to pre-compute all augmentations before training. Otherwise, it is better to sample the augmentations for every patch and epoch before training, pre-computing just those combinations.}. As a result of this overhead, practitioners must carefully choose which augmentations to apply, if any. We address this problem by assessing the performance benefit obtained by different augmentations on our benchmark tasks.

\section{Related work}
\subsection{Weakly supervised \gls{wsi} classification}
Early work on \gls{wsi} classification with slide-level labels employed \glspl{cnn} such as ResNet~\cite{he2015deep} which were pretrained on ImageNet~\cite{deng2009imagenet} and then fine-tuned on the classification task using slide-level labels as patch-level supervision~\cite{coudray2018classification,kather2019deep}.
Recognising that this approach introduces excessive noise in the patch-level supervision to the detriment of the training process, later work~\cite{ilse2018attention,wang2018revisiting} reframed this task as an embedding-based \gls{mil} problem~\cite{dietterich1997solving}. 
In this line of work, a feature vector is extracted for every patch using a \gls{cnn} ($f$ in \cref{fig:overview}), and these feature vectors are aggregated and classified via a learnable pooling function and classifier ($h_\theta\circ g_\theta$ in \cref{fig:overview}). 
Initially, the entire network, including feature extraction, was trained end-to-end~\cite{ilse2018attention}. 
However, end-to-end training becomes intractable as \gls{mil} approaches scale to larger datasets, so more recent models operate on frozen features extracted using ImageNet pretrained models~\cite{lu2021dataefficient}. 
The frozen feature approach is now widely adopted for weakly supervised learning on \glspl{wsi}, albeit with better feature extractors trained using \gls{ssl}.

\subsection{\Gls{ssl} in pathology}
The goal of \gls{ssl} is to learn useful representations for downstream tasks from unlabelled data. 
Unlike supervised learning, \gls{ssl} leverages structures inherent in data through pretext tasks, without needing explicit labels. 
The development of \gls{ssl} models is an active area of research, from which a variety of algorithms like contrastive learning~\cite{chen2020simple,he2020momentum,chen2021empirical}, non-contrastive learning~\cite{grill2020bootstrap,zbontar2021barlow} and clustering-based methods~\cite{caron2018deep,caron2020unsupervised} have emerged in recent years, each with unique advantages and challenges. 
These models have quickly found adoption in the pathology field, which is well-situated to benefit from \gls{ssl} due to the availability of large datasets that lack patch-level labels. 
Indeed, \gls{ssl} feature extractors pretrained on pathology data have been shown to outperform ImageNet pretrained models on downstream pathology tasks~\cite{shao2023generalizability,kang2023benchmarking,campanella2023computational,chen2021selfsupervisedhistopathology,sikaroudi2023generalization}. 

In the last three years, a number of models have been developed~\cite{wang2021transpath,chen2022scaling,wang2022transformer,chen2024uni,azizi2023robust,wang2023retccl,filiot2023scaling,kang2023benchmarking,vorontsov2023virchow,lu2024conch,lazard2023giga} that were pretrained using \gls{ssl} on large multi-centre pathology datasets, such as \gls{tcga}~\cite{weinstein2013cancer}.
Wang \etal~\cite{wang2022transformer,wang2021transpath} proposed CTransPath, a Swin Transformer~\cite{liu2021swin} feature extractor trained using \gls{srcl}, a novel \gls{ssl} technique based on MoCo~v3~\cite{chen2021empirical} specifically tailored to pathology. 
Previously, they had put forth RetCCL~\cite{wang2023retccl}, a ResNet-50 model trained using a \gls{ssl} technique they termed \gls{ccl} based on MoCo~\cite{he2020momentum}. 
Lunit~\cite{kang2023benchmarking} benchmarked four \gls{ssl} techniques, Barlow Twins~\cite{zbontar2021barlow}, SwAV~\cite{caron2020unsupervised}, MoCo~v2~\cite{chen2020improved}, and DINO~\cite{caron2021emerging}, for pathology by training them on \gls{tcga}.
Owkin~\cite{filiot2023scaling} evaluated different ViT variants~\cite{kolesnikov2021image} using the iBOT framework~\cite{zhou2022image}, terming their best ViT-B variant ``Phikon''.
All eight aforementioned models are available publicly, and form the basis of our study (we include both the student and teacher variants of Phikon).
We refer the reader to \cref{sec:app:feature_extractors} for a more detailed overview.

Trained on even larger datasets, a number of closed-source pathology foundation models have emerged in the last year~\cite{chen2024uni,vorontsov2023virchow,campanella2023computational,azizi2023robust,lu2024conch,dippel2024rudolfv}.
As a notable exception that was made available to the research community, UNI~\cite{chen2024uni} was trained on over 100,000 slides using DINOv2~\cite{oquab2023dinov2}.
Whilst UNI is the only foundation model we could include in our study, we provide a more detailed account of the other closed-weights pathology foundation models in \cref{sec:app:foundation_models}.

\subsection{Stain normalisation}
Different medical sites employ different microscopes, scanners, protocols, and dyes, resulting in variations in \glspl{wsi} appearance.
For over 20 years~\cite{ruifrok2001quantification,reinhard2001color,macenko2009method}, stain normalisation has been commonplace in digital pathology workflows to account for these factors by adjusting colours to match a reference image.
Classical techniques~\cite{reinhard2001color,macenko2009method,vahadane2016structure} achieve this by performing colour deconvolution, standardising stain intensity, and then transforming the colour space of the input images to that of a reference image.
More recently, GAN-based approaches have been proposed to this end as well~\cite{wagner2021structure,nazki2023multipathgan,zanjani2018stain}.
Boschman \etal~\cite{boschman2022utility} compared eight classical and GAN-based stain normalisation techniques, concluding that stain normalisation, especially Vahadane~\cite{vahadane2016structure} and Macenko~\cite{macenko2009method} normalisation, can indeed bolster slide-level classification performance when validating on external datasets.
However, their approach aggregated patch-level predictions via simplistic majority vote and did not integrate \gls{ssl} feature extractors.
In contrast, we contend that with \gls{ssl} feature extractors, stain normalisation becomes obsolete.
To show this, we focus our analysis on Macenko normalisation~\cite{macenko2009method}, the technique most widely adopted in the literature~\cite{ghaffari2022benchmarking,chikontwe2022weakly,schrammen2022weakly,elnahhas2024regressionbased,elnahhas2023whole,liu2023identification}. 

\subsection{Image augmentations}
As a common regularisation technique for neural network training in general~\cite{cubuk2021tradeoffsiclr}, image augmentations and have unsurprisingly found widespread adoption in histopathology as well~\cite{salvi2021impact}.
In this field, the most popular augmentations include flipping, scaling, rotating, and colour alterations due to the nature of pathology slides~\cite{salvi2021impact}, though a recent line of research introduces ``stain augmentation'' as a combination of stain normalisation and image augmentations to increase data diversity as well~\cite{tellez2018hande,marini2023datadriven,shen2022randstainna}.
In this work, we study 26 image augmentations, focusing our analysis on those popular in pathology. 

\subsection{Robustness of feature extractors in pathology}
Assessing the robustness and generalisation ability of deep learning pathology models in the face of domain shift and \gls{ood} data is an active area of research~\cite{jahanifar2023domain,zhang2022benchmarking,schomig-markiefka2021quality,ghaffarilaleh2022adversarial} and an important undertaking, considering the stakes may be human life.
Our work builds upon Lunit's aforementioned \gls{ssl} benchmarking initiative~\cite{kang2023benchmarking}, which involves training and evaluating four pathology-oriented \gls{ssl} feature extractors which we include in our study.
Lunit's evaluation, however, is confined to patch classification and nuclei segmentation.
While such tile-based tasks are scientifically interesting and the predominant means of evaluation in the literature~\cite{tellez2019quantifying,kang2023benchmarking,springenberg2023modern}, it has been suggested~\cite{campanella2023computational} that for evaluations to have greater clinical relevance, they should instead focus on slide-level tasks -- predicting patient variables such as prognostic outcomes and biomarkers -- and validate results on independent external cohorts.
In response to this, we evaluate a total of eight \gls{ssl} feature extractors across nine slide-level classification targets (whose clinical utility we detail in \cref{app:downstream_tasks}), using external cohorts that were unseen during training (both in \gls{ssl} pretraining and downstream evaluation).

Similar to our work, Tellez \etal~\cite{tellez2019quantifying} explore the influence of stain normalisation and image augmentations on the generalisability of pathology models.
However, their 2019 study predates \gls{ssl} models trained on expansive pathology datasets akin to those employed in our evaluation; their analysis is limited to \gls{cnn}s trained from scratch on narrow patch classification tasks.
Springenberg \etal~\cite{springenberg2023modern} empirically assess the robustness of \gls{cnn}s and ViTs in pathology with and without \gls{ssl} pretraining (CTransPath~\cite{wang2022transformer} and RetCCL~\cite{wang2023retccl}), but their evaluation, again, is confined to patch classification.
Sikaroudi \etal~\cite{sikaroudi2023generalization} compare the \gls{ood} generalisability of pathology pretrained models (focusing on supervised and self-supervised models trained on natural images as well as a non-\gls{ssl} pathology-specific model~\cite{riasatian2021fine}), but also only consider patch classification. 

\section{Effect on latent space}
\label{sec:latent}
\begin{figure}
    \centering
    \includegraphics[width=\linewidth]{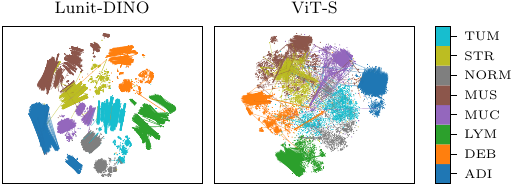}
    \caption{Latent space visualisations (t-SNE~\cite{vandermaaten2008visualizing}) of features extracted with Lunit-DINO (left) \vs ImageNet baseline (right). Colours represent tissue classes~\cite{kather2018100000}. Both feature extractors use the same architecture (ViT-S), but the left was trained on pathology images using \gls{ssl}. Each dot represents a feature vector in latent space extracted from an unaltered image patch, and we draw a line from that dot to the corresponding stain-normalised version.}
    \label{fig:latent_macenko}
\end{figure}
An ideal feature extractor for pathology extracts meaningful features from a patch. More specifically, it should
(i) be invariant to factors we deem unimportant, \eg stain, orientation, \etc, and
(ii) vary with properties we are interested in, \eg tissue type, cell type, and many other factors not known \emph{a priori}.
For example, a good feature extractor will produce a similar embedding for a particular patch and its stain-normalised version (as we want the feature extractor to be invariant to this factor), but yield very different embeddings for two patches of different tissue classes (\ie normal \vs tumour). 

In this section, we study the effect of various augmentations on the latent space, beginning with stain normalisation. 
We employ the NCT-CRC-HE-100K dataset~\cite{kather2018100000,kather2019predicting}, comprising 100,000 patches extracted from \gls{he} \gls{crc} images without stain normalisation.
This dataset includes patch-level tissue type labels which enables more fine-grained analysis and visualisation. 

\subsection{Stain normalisation}
\label{sec:latent_normalisation}
How similar are feature vectors extracted from image patches to those derived from their stain-normalised counterparts? 
We contend that simply looking at the average distance between original embeddings and their stain-normalised versions does not provide enough information to make claims about the quality of a feature extractor. 
To obtain a more nuanced view of how stain normalisation affects embeddings, we present a dimensionality-reduced latent space visualisation of Lunit's DINO feature extractor in \cref{fig:latent_macenko}. 
This feature extractor is highlighted due to its superior downstream performance (see \cref{fig:feature_extractor_performance_comparison} and analysis in \cref{sec:feature_extractor_comparison}). 
In our visualisation, each point corresponds to a feature vector, with a line connecting each original feature vector to its stain-normalised version. 
Notably, Lunit-DINO clusters tissue types in latent space and the displacement of the feature vectors induced by stain normalisation is largely confined to these clusters. 
In contrast, a baseline extractor using the same ViT-S architecture~\cite{liu2021swin} but trained via supervised learning on ImageNet, demonstrates less effective clustering and exhibits a different pattern: some features move hardly at all while others make large jumps between clusters, as indicated by the longer inter-cluster lines in \cref{fig:latent_macenko}, right. 
In fact, this pattern is consistent across various feature extractors: those pretrained on pathology data are less prone to `jump' between tissue type clusters compared to their ImageNet-pretrained counterparts when undergoing stain normalisation, further detailed in \cref{sec:app:augmentations}.

\begin{figure}
    \centering
    \includegraphics[width=\linewidth]{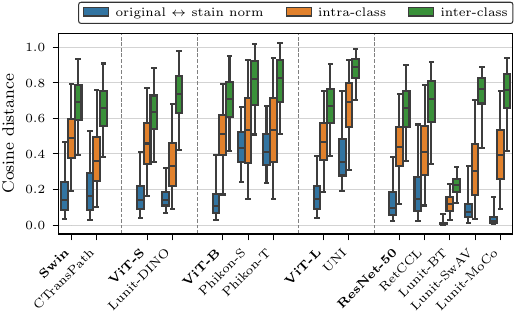}
    \caption{Boxplot of cosine distances between patch embeddings and their \textcolor{figblue}{stain-normalised versions}, as well as between embeddings of randomly chosen patches \textcolor{figorange}{of the same} or \textcolor{figgreen}{of differing} tissue types. Feature extractors are grouped by architecture (ImageNet baselines are \textbf{bold}). Whiskers represent 95\% of the distances.}
    \label{fig:macenko_feature_extractor_comparison}
\end{figure}

In \cref{fig:macenko_feature_extractor_comparison}, we compare the cosine distances of the embedding displacement caused by stain normalisation across all 14 feature extractors. 
Despite the important difference in terms of intra-cluster \vs inter-cluster jumps identified in the latent space visualisation above, Lunit-DINO and ViT-S exhibit similar averages (\cf their medians in \cref{fig:macenko_feature_extractor_comparison}, \textcolor{figblue}{blue}). 
This observation highlights the importance of examining the distribution of distances, not merely their averages: the boxplot in \cref{fig:macenko_feature_extractor_comparison} reflects this difference by the increased range of the whiskers of ViT-S compared to Lunit-DINO. 

We note that an analysis that considers embedding displacement only from the perspective of stain normalisation is insufficient to make meaningful claims about feature extractor utility. 
For example, an extractor that maps all images to a single point in latent space would negate any embedding displacement induced by stain normalisation and prevent inter-cluster jumps, yet its features would be wholly useless to the downstream model. 
This observation leads us to also consider the second key criterion outlined at the beginning of this section: the ability of feature extractors to vary embeddings according to characteristics critical for downstream tasks. 
We select tissue type a surrogate marker to investigate this second criterion. 
However, it is important to recognise as a limitation of this analysis that there are numerous other potentially significant characteristics that remain unidentified at this stage, and for which specific labels are unavailable. 
Nonetheless, we posit that feature vectors from similar tissue types (indicated in \textcolor{figblue}{blue} in \cref{fig:macenko_feature_extractor_comparison}) should be closer in latent space compared to those from different tissue types (shown in \textcolor{figgreen}{green}). 
Upon examining the disparity between these distance measures, we find that the ImageNet baselines tend to lump all features more closely together, regardless of tissue type. 
In contrast, the \gls{ssl} models show better differentiation, as indicated by a greater separation between the blue and green boxes in the boxplot. 
Furthermore, the extent to which patches of different tissue types are distanced in the latent space  (\textcolor{figgreen}{green}) also provides a useful scale for contextualising the original \vs stain-normalised distances (\textcolor{figblue}{blue}). 
These findings suggest that the choice of pretraining data influences the stability of feature vectors in the context of stain normalisation. 
More specifically, feature extractors that have seen diverse stains as part of their \gls{ssl} pretraining can learn to become more robust to variations in stain, while still preserving variations in aspects relevant to downstream tasks, \ie tissue type.

\subsection{Image augmentations}
\label{sec:latent_augmentations}

\begin{figure}
    \centering
    \includegraphics[width=\linewidth]{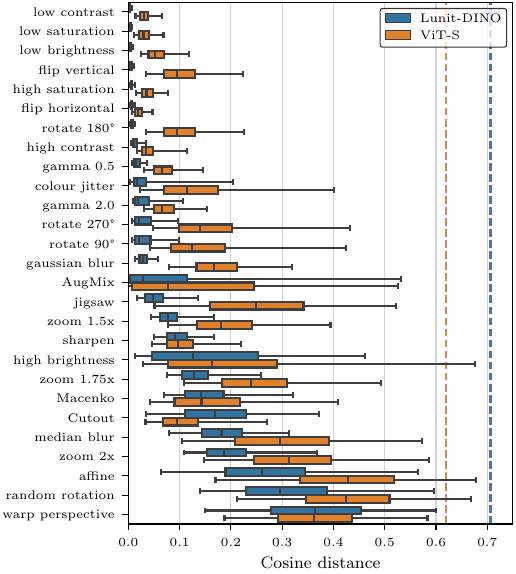}
    \caption{Boxplot of embedding displacement induced by image augmentations for \textcolor{figblue}{Lunit-DINO} and \textcolor{figorange}{ViT-S}. Dashed lines represent the average distance between randomly selected patches (without augmentation) indicating how `dispersed' the latent spaces are.}
    \label{fig:embedding_displacement_by_augmentation}
\end{figure}

In principle, the methodology presented above is suitable to study how \emph{any} transformation of input patches, not just stain normalisation, manifests itself in latent space. 
Here, we consider 26 common image augmentations, for which we provide representative examples in \cref{sec:app:augmentations}. 
For Lunit-DINO and the ViT-S baseline, we compare the magnitudes of the embedding displacement across augmentations in \cref{fig:embedding_displacement_by_augmentation}. 
We observe that Lunit-DINO's embeddings are more robust to augmentations compared to the ImageNet baseline: for all augmentations except `Cutout'~\cite{devries2017improved} and `warp perspective', the cosine distances tend to be smaller in Lunit-DINO. 
That is even though Lunit-DINO's embeddings are generally more spread out (the average distance between any two randomly selected non-augmented patches is greater, indicated by the dashed lines in \cref{fig:embedding_displacement_by_augmentation}). 
Normalising the distances by this average, Cutout remains the only (minor) exception. 

\begin{figure}
  \centering
  \includegraphics[width=\linewidth]{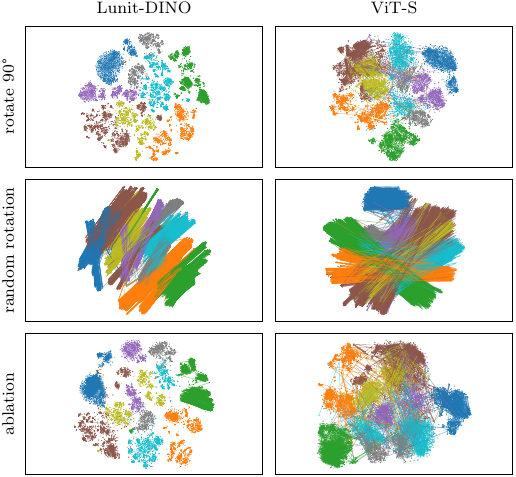}
  \caption{Visualisations of latent space transformations caused by rotation-based augmentations (rows) in Lunit-DINO (left) and ViT-S (right). Colours and lines are as explained in \cref{fig:latent_macenko}. Top row: 90° rotation. Middle: rotating by a random angle. Bottom (ablation): each line represents the transformation from (a) the embedding of the $1.5\times$ zoomed version of a patch to (b) the embedding obtained by randomly rotating before the $1.5\times$ zoom.}
  \label{fig:latent_augmentations}
\end{figure}

We observe that Lunit-DINO excels in terms of robustness to right-angle rotations and flips -- a much desired property considering that \glspl{wsi}, unlike natural images, lack a canonical orientation. 
In fact, in selecting augmentations for generating positive pairs for \gls{ssl} pretraining of the Lunit feature extractors, Kang \etal~\cite{kang2023benchmarking} employed the aforementioned augmentations for this precise reason, incentivising rotated/flipped embeddings to be close in latent space. 
On the other hand, the ImageNet baseline is significantly less robust to such augmentations. 
Interestingly, it is more robust to horizontal flips than vertical flips, which may be explained by the fact that it was trained on natural images.

Although Lunit-DINO is remarkably robust to rotating by angles that are multiples of 90°, non-right angles cause the greatest displacement in latent space aside from perspective warp (penultimate row in \cref{fig:embedding_displacement_by_augmentation}).
To investigate this, we visualise the latent space in \cref{fig:latent_augmentations}. 
As expected, for 90° rotation (top row), Lunit-DINO's latent space remains largely unchanged, as opposed to ViT-S. 
However, for random rotations (middle row), we observe a high degree of chaotic jumps in both feature extractors, indicating neither is robust to this augmentation. 
We hypothesise that this is caused by the loss of pixels at the edges of patches in off-angle rotations, and design an ablation study to investigate this phenomenon. 
To eliminate the black pixel problem, we perform a centercrop on the original and augmented patches in a manner that ensures there are no black pixels in any rotation. 
The corresponding latent space visualisations at the bottom of \cref{fig:latent_augmentations} confirm our assumption: Lunit-DINO's latent space remains unchanged whereas ViT-S's embeddings move significantly. 
Similar reasoning may explain the poor robustness regarding `random affine', `warp perspective', and `Cutout'~\cite{devries2017improved}.

\section{Impact on downstream performance}
\label{sec:downstream}

Motivated by the findings above, specifically that some augmentations have larger effects than others on the latent representations, we investigate in the remainder of this paper how stain normalisation and augmentations affect downstream performance. 
To do so, we train weakly supervised models on nine downstream tasks using publicly available datasets. 

\subsection{Experimental setup}
\subsubsection{Models}
We compare three parametrisations of the downstream aggregation model $g_\theta(\cdot)$ in \cref{eq:mil_aug}:
  (1) AttMIL~\cite{ilse2018attention}, the most common approach in the literature,
  (2) a two-layer decoder-only transformer \cite{vaswani2017attention}, which is gaining popularity in recent works~\cite{wagner2023transformerbased,wolflein2023deep}, and
  (3) a simple baseline performing mean average pooling across features as
  \setlength{\abovedisplayskip}{2pt}
  \setlength{\belowdisplayskip}{2pt}
  \begin{equation*}
    g_\theta\left(\{x_i\}_{i=1}^n\right) = \frac{1}{n} \sum_{i=1}^n{x_i}
  \end{equation*}
  where each $x_i$ is a feature vector.
In all experiments, we parametrise $h_\theta(\cdot)$ as a linear layer with a softmax over the number of classes for the given task. 

\subsubsection{Tasks and datasets}
In selecting downstream tasks, we prioritise those with clinical utility and whose underlying variables are also available in adequately sized public datasets. 
Training on \gls{tcga}-BRCA~\cite{weinstein2013cancer} and testing on CPTAC-BRCA~\cite{krug2020proteogenomic}, we predict four breast cancer (\breasticon) targets: subtype as well as the CDH1, TP53, and PIK3CA genetic mutations. 
Furthermore, we predict \breasticon-lymph node status in the CAMELYON17 breast cancer dataset~\cite{bandi2018detection} (which contains data from five centres -- we used one of the centres for testing and the others for training).
Finally, we predict four markers in colorectal cancer (\colonicon): \acs{msi} status as well as BRAF, KRAS, and SMAD4 genetic mutations (training on \gls{tcga}-CRC~\cite{weinstein2013cancer} and testing on CPTAC-COAD~\cite{vasaikar2019proteogenomic}).
We elaborate on these variables, their clinical relevancy, and the underlying datasets in \cref{app:downstream_tasks}. 

Our choice of test datasets is deliberate:
we employ external cohorts for testing to assess generalisability on unseen datasets, and
ensure that the test datasets are non-overlapping with the \gls{ssl} pretraining datasets to avoid data leakage from the feature extractors.

\subsubsection{Training details}
We train each model using the AdamW optimiser~\cite{loshchilov2017decoupled} with an initial learning rate of 0.001 which is decayed using cosine annealing~\cite{loshchilov2016sgdr} for up to 30 epochs, though training typically ends sooner due to our use of early stopping (when the validation loss fails to improve for ten consecutive epochs).
For this, we allocate 80\% of the training set for model training and 20\% for validation.
We conduct training with five distinct random seeds for the cartesian product of the 14 feature extractors, nine tasks, three downstream models, and six preprocessing/augmentation setups (slidewise or patchwise stain normalisation, rotate/flip, all augmentations, or none), resulting in over 8,000 trained models. 
The training and validation splits are kept fixed per-task across the seeds for all tasks except for lymph node status classification. 
This latter task uses the CAMELYON17 dataset, allowing us to perform leave-one-hospital-out cross-validation with a different random seed for each of the five hospitals. 
For the experiments involving augmentations, we apply these augmentations only on the images of the training datasets, never the test datasets (except for the stain normalisation experiments, where we ensure the same normalisation is applied to training and test datasets). 
We perform feature extraction once before training, caching for every patch in every dataset its original feature vector as well as the feature vectors of all 27 augmented versions of that patch, including stain normalisation. 
More details are provided in \cref{sec:app:caching}.

\subsection{Lunit-DINO, UNI, and CTransPath extract the most useful features}
\label{sec:feature_extractor_comparison}

\begin{figure}[t]
  \centering
  \includegraphics[width=\linewidth]{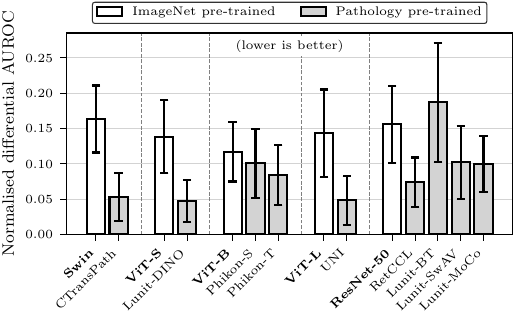}
  \caption{%
    Task-averaged relative performance comparison of the evaluated feature extractors, using the AttMIL aggregation model.
    For each task and feature extractor, we train five models with different random seeds.
    The relative performance is calculated as the normalised differential \acs{auroc} score (lower is better), which reflects the expected decrease in test \acs{auroc} when selecting a given feature extractor relative to the best-performing one, considering every way of choosing one random seed per feature extractor (\cf \cref{sec:feature_extractor_comparison}).
    We show the task-averaged performance here, which corresponds to the last column in \cref{tbl:norm_differential_auroc_attmil_noaug}.
  }
  \label{fig:feature_extractor_performance_comparison}
\end{figure}

Having trained a large number of downstream models based on 14 feature extractors across a diverse set of tasks, we are in a position to identify the most effective feature extractor overall.
We present these findings first and focus our later discussion on these feature extractors in particular.

\begin{table*}[t]
  \caption{Comparative evaluation of feature extractors. This table presents the normalised differential \acs{auroc} scores (lower is better) for all feature extractors, across the evaluated targets using the AttMIL~\cite{ilse2018attention} aggregation model. The scores reflect the expected decrease in test \acs{auroc} when selecting a given feature extractor relative to the best-performing one (\cf \cref{sec:feature_extractor_comparison}).}
  \label{tbl:norm_differential_auroc_attmil_noaug}
  \resizebox{\textwidth}{!}{%
      \begin{tabular}{l|ccccccccc|c}
          \toprule
          Feature extractor & \breasticon-subtype & \breasticon-CDH1 & \breasticon-TP53 & \breasticon-PIK3CA & \breasticon-LN status & \colonicon-\acs{msi} & \colonicon-KRAS & \colonicon-BRAF & \colonicon-SMAD4 & Average \\
          \midrule
          Swin~\cite{liu2021swin} & $0.10 \pm 0.02$ & $0.17 \pm 0.02$ & $0.31 \pm 0.02$ & $0.07 \pm 0.04$ & $0.20 \pm 0.09$ & $0.18 \pm 0.04$ & $0.14 \pm 0.04$ & $0.14 \pm 0.06$ & $0.16 \pm 0.05$ & $0.163 \pm 0.047$ \\
          CTransPath~\cite{wang2022transformer} & $0.03 \pm 0.02$ & $\mathbf{0.01 \pm 0.01}$ & $0.05 \pm 0.02$ & $0.05 \pm 0.03$ & $0.08 \pm 0.07$ & $0.09 \pm 0.03$ & $0.06 \pm 0.03$ & $0.06 \pm 0.03$ & $0.06 \pm 0.03$ & $\underline{0.053 \pm 0.034}$ \\
          ViT-S~\cite{kolesnikov2021image} & $0.16 \pm 0.02$ & $0.08 \pm 0.03$ & $0.17 \pm 0.05$ & $0.08 \pm 0.05$ & $0.22 \pm 0.09$ & $0.19 \pm 0.04$ & $0.06 \pm 0.03$ & $0.20 \pm 0.04$ & $0.08 \pm 0.08$ & $0.138 \pm 0.052$ \\
          Lunit-DINO~\cite{kang2023benchmarking} & $0.10 \pm 0.02$ & $0.04 \pm 0.03$ & $0.06 \pm 0.03$ & $0.02 \pm 0.03$ & $0.09 \pm 0.04$ & $\mathbf{0.01 \pm 0.01}$ & $0.06 \pm 0.04$ & $\mathbf{0.02 \pm 0.03}$ & $0.02 \pm 0.02$ & $\mathbf{0.047 \pm 0.030}$ \\
          ViT-B~\cite{kolesnikov2021image} & $0.11 \pm 0.04$ & $0.11 \pm 0.02$ & $0.19 \pm 0.03$ & $0.08 \pm 0.03$ & $0.20 \pm 0.06$ & $0.15 \pm 0.03$ & $0.03 \pm 0.04$ & $0.18 \pm 0.07$ & $\mathbf{0.01 \pm 0.01}$ & $0.117 \pm 0.042$ \\
          Phikon-S~\cite{filiot2023scaling} & $0.12 \pm 0.01$ & $0.09 \pm 0.02$ & $0.13 \pm 0.03$ & $0.09 \pm 0.03$ & $0.09 \pm 0.08$ & $0.07 \pm 0.04$ & $0.07 \pm 0.04$ & $0.07 \pm 0.06$ & $0.17 \pm 0.08$ & $0.100 \pm 0.049$ \\
          Phikon-T~\cite{filiot2023scaling} & $0.09 \pm 0.02$ & $0.10 \pm 0.03$ & $0.10 \pm 0.05$ & $0.08 \pm 0.03$ & $0.06 \pm 0.04$ & $0.05 \pm 0.04$ & $0.08 \pm 0.04$ & $0.09 \pm 0.07$ & $0.09 \pm 0.04$ & $0.084 \pm 0.043$ \\
          ViT-L~\cite{kolesnikov2021image} & $0.15 \pm 0.03$ & $0.03 \pm 0.02$ & $0.18 \pm 0.04$ & $0.06 \pm 0.05$ & $0.18 \pm 0.08$ & $0.19 \pm 0.11$ & $0.15 \pm 0.06$ & $0.16 \pm 0.06$ & $0.19 \pm 0.06$ & $0.143 \pm 0.062$ \\
          UNI~\cite{chen2024uni} & $\mathbf{0.00 \pm 0.01}$ & $0.06 \pm 0.04$ & $\mathbf{0.00 \pm 0.00}$ & $0.06 \pm 0.03$ & $\mathbf{0.01 \pm 0.02}$ & $0.04 \pm 0.04$ & $0.04 \pm 0.02$ & $0.13 \pm 0.07$ & $0.10 \pm 0.04$ & $\underline{0.048 \pm 0.034}$ \\
          ResNet-50~\cite{he2015deep} & $0.17 \pm 0.02$ & $0.09 \pm 0.04$ & $0.14 \pm 0.03$ & $\mathbf{0.02 \pm 0.02}$ & $0.20 \pm 0.08$ & $0.23 \pm 0.04$ & $0.11 \pm 0.03$ & $0.23 \pm 0.07$ & $0.21 \pm 0.09$ & $0.156 \pm 0.054$ \\
          RetCCL~\cite{wang2023retccl} & $0.09 \pm 0.03$ & $0.04 \pm 0.02$ & $0.07 \pm 0.03$ & $0.05 \pm 0.03$ & $0.10 \pm 0.07$ & $0.09 \pm 0.03$ & $\mathbf{0.03 \pm 0.02}$ & $0.14 \pm 0.03$ & $0.06 \pm 0.03$ & $0.074 \pm 0.035$ \\
          Lunit-BT~\cite{kang2023benchmarking} & $0.15 \pm 0.03$ & $0.07 \pm 0.03$ & $0.05 \pm 0.01$ & $0.13 \pm 0.04$ & $0.36 \pm 0.15$ & $0.28 \pm 0.13$ & $0.03 \pm 0.04$ & $0.35 \pm 0.13$ & $0.25 \pm 0.03$ & $0.187 \pm 0.085$ \\
          Lunit-SwAV~\cite{kang2023benchmarking} & $0.09 \pm 0.02$ & $0.07 \pm 0.03$ & $0.09 \pm 0.02$ & $0.13 \pm 0.06$ & $0.10 \pm 0.06$ & $0.11 \pm 0.03$ & $0.13 \pm 0.06$ & $0.07 \pm 0.07$ & $0.14 \pm 0.08$ & $0.102 \pm 0.052$ \\
          Lunit-MoCo~\cite{kang2023benchmarking} & $0.10 \pm 0.04$ & $0.08 \pm 0.02$ & $0.07 \pm 0.02$ & $0.07 \pm 0.03$ & $0.11 \pm 0.06$ & $0.20 \pm 0.06$ & $0.08 \pm 0.05$ & $0.11 \pm 0.03$ & $0.07 \pm 0.03$ & $0.099 \pm 0.040$ \\
          \bottomrule
      \end{tabular}
  }\vspace{-3mm}
\end{table*}

First, let us consider how to determine the best feature extractor for a given task and downstream aggregator (such as predicting \breasticon-CDH1 with AttMIL aggregation).
For any such task-model pair, we trained 70 models -- spanning the 14 feature extractors across five random seeds.
We define a `trial' as one particular configuration pairing each feature extractor with a random seed, leading to $5^{14}$ ($\approx$ 6 billion) unique trials.
Within each trial, we evaluate the feature extractors based on the difference between their test \gls{auroc} and the highest test \gls{auroc} observed in that trial, thus assigning a score of zero to the top performer.
By calculating the mean across all $5^{14}$ trials, we derive the `normalised differential \gls{auroc} score' -- a measure that captures the relative efficacy of the feature extractors and allows fair comparisons across tasks of varying difficulty.

More formally, let $A_{t,m,f_i,s_j}\in\mathbb{R}$ denote the test \gls{auroc} of the model trained on task $t$ with feature extractor $f_i$ and downstream aggregation model $m$, using random seed $s_j\in\mathcal{S}$ (the set of seeds $\mathcal{S}=\{1,\dots,5\}$).
Here, $i\in\{1,\dots,F\}$ indexes the feature extractors, and $F=14$ is the total number of feature extractors in our study.
Then, the normalised differential \gls{auroc} score for a given task $t$, downstream model $m$, and feature extractor $f_i$ is defined as
\begin{multline}
  \label{eq:norm_diff_auroc}
  \text{NDS}_{t,m,f_i} \\
  = \mathbb{E}_{(s_1,\dots,s_F)\in\mathcal{S}^F}
  \left[\max_{j=1,\dots,F} A_{t,m,f_j,s_j} - A_{t,m,f_i,s_i}\right],
\end{multline}
where $\mathcal{S}^F$ is the Cartesian product of $\mathcal{S}$ taken $F$ times.
Thus, the expectation is taken over all possible permutations of choosing $F$ random seeds $s_1,\dots,s_F$, one for each feature extractor.
This means that for a particular downstream task and aggregation model, the normalised differential AUROC represents the expected decrease in test \gls{auroc} when selecting a given feature extractor $f_i$ relative to the best-performing one.

The outcomes of this analysis, when considering the downstream AttMIL model and no augmentations, are detailed individually per task in \cref{tbl:norm_differential_auroc_attmil_noaug} and averaged across tasks in \cref{fig:feature_extractor_performance_comparison}.
Notably, Lunit-DINO, UNI, and CTransPath achieve the best task-averaged performance.
Indeed, they consistently perform best, regardless of downstream aggregation model and type of input augmentation, as we show in the extended data table in \cref{sec:app:extended_data_and_figures}.
Unsurprisingly, we find the ImageNet baselines perform worse than the pathology models (with the exception of Lunit-BT which performs very poorly indeed), which is in line with previous work~\cite{dehaene2020selfsupervision,ciga2022self,kang2023benchmarking,campanella2023computational,chen2021selfsupervisedhistopathology,sikaroudi2023generalization}.

\subsection{Stain normalisation does not impact downstream performance}
\label{sec:stain_normalisation_downstream}

\begin{figure*}[t]
  \centering
  \includegraphics[width=\linewidth]{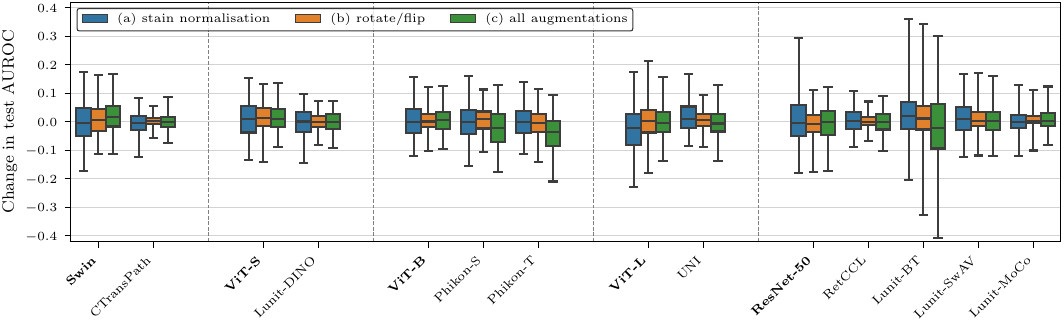}
  \caption{%
      \textbf{Stain normalisation and image augmentations do not meaningfully impact downstream performance.}
      The boxplots show the expected difference in test \acs{auroc} between models trained (a) \textcolor{figblue}{with \vs without stain normalisation}, (b) \textcolor{figorange}{rotation/reflection augmentations \vs no augmentations}, (c) \textcolor{figgreen}{all augmentations \vs no augmentations} for each feature extractor.
      We observe no benefit in employing stain normalisation or augmentations before feature extraction. 
      The best feature extractors, Lunit-DINO, UNI, and CTransPath (\cf \cref{fig:feature_extractor_performance_comparison}), are particularly robust, unlike ImageNet baselines (\textbf{bold}).
    }
  \label{fig:augmentations_performance_effect}
  \vspace{-10pt}
\end{figure*}

We quantify the effect of stain normalisation on downstream model performance by determining the expected difference in test \gls{auroc} between models trained with stain normalisation \vs without.
Given a feature extractor and downstream aggregation model, \eg Lunit-DINO with AttMIL, we must compare 45 models trained with stain normalisation (nine tasks times five random seeds) with another 45 models trained without stain normalisation. 
To estimate the difference in \gls{auroc}, we perform bootstrapping\footnote{This is not to be confused with the normalised differential \gls{auroc} score from the previous section, which is a measure of relative performance and does not involve bootstrapping.}: 
For each of the 45 task-seed pairs, we generate 25 random resamples of the respective test datasets with replacement, totalling $45 \times 25 = 1{,}125$ bootstraps. 
Since each bootstrap is associated with a particular task-seed combination, it corresponds to two trained models: one trained with stain normalisation and one without. 
We deploy both models on the given bootstrap, computing the difference in \gls{auroc}. 
Repeating for all bootstraps, we obtain a distribution of 1,125 \gls{auroc} differences which we present as a boxplot in \cref{fig:augmentations_performance_effect}, with a separate box for every feature extractor (we focus on the AttMIL~\cite{ilse2018attention} aggregation model because it is the most widely used, but provide analogous plots for the other two in \cref{sec:app:extended_data_and_figures}). 
We find no clear \gls{auroc} difference between the two groups, for any feature extractor: all 95\% confidence intervals (and interquartile ranges) include zero. 
Surprisingly, this holds even for ImageNet extractors, whose latent spaces we previously identified more susceptible to larger displacements due to stain normalisation. 

\subsubsection{Slidewise versus patchwise normalisation}
While in \cref{fig:augmentations_performance_effect}a, we perform stain normalisation on a per-slide basis, a more computationally efficient%
\footnote{%
  Macenko normalisation~\cite{macenko2009method} requires an eigenvalue decomposition across all pixels, scaling cubically in the number of pixels. 
  For a slide with $n$ patches of $k$ pixels each, the complexity is in $\mathcal{O}(n^3k^3)$ for slidewise normalisation, but $\mathcal{O}(nk^3)$ for patchwise. 
  Moreover, the latter is embarrassingly parallel across the $n$ dimension.
} alternative is normalising each patch individually.
However, in this approach, adjacent patches might experience different colour transformations, potentially affecting consistency across the slide. 
We perform an ablation study, detailed in \cref{sec:app:stain_normalisation}, where we employ the bootstrapping procedure from above with patchwise instead of slidewise normalisation, but find no consistent performance differences between both methods. 
Therefore, we recommend the patchwise approach for practitioners still seeking to employ stain normalisation in their preprocessing pipelines, due to its computational benefit. 

\subsection{Augmentations do not impact performance}
Having emphasised rotation and reflection as augmentations of particular relevance to pathology in our investigation of the latent space, we now study their downstream impact on performance.
To this end, we trained a batch of models where at each epoch, each patch is randomly flipped (horizontally or vertically) or rotated by a right angle before feature extraction.
Analogous to our analysis of stain normalisation (\cref{fig:augmentations_performance_effect}a), we perform a bootstrapped quantification of the difference in performance incurred by employing the augmented versus non-augmented features in \cref{fig:augmentations_performance_effect}b.
Again, we observe no consistent benefit in employing this type of augmentation.
Furthermore, expanding the set of augmentations to all 27 studied transformations (with each being equally likely to be selected for every patch at every epoch) yields similar results (\cref{fig:augmentations_performance_effect}c).
While \cref{fig:augmentations_performance_effect} employs AttMIL~\cite{ilse2018attention}, we come to the same conclusion for the other downstream aggregation architectures, for which we present extended results in \cref{sec:app:extended_data_and_figures}. 

\subsection{Downstream aggregation models}
\begin{figure}
  \centering
  \includegraphics[width=0.5\textwidth]{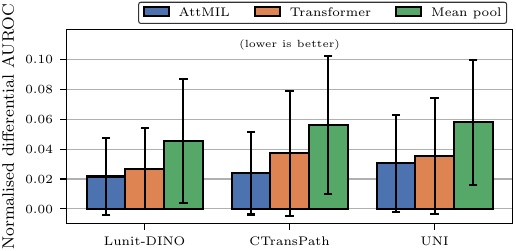}
  \caption{Performance impact of choosing a particular downstream aggregation model.}
  \label{fig:norm_diff_auroc_downstream_model}
\end{figure}
In \cref{sec:feature_extractor_comparison}, we identified Lunit-DINO, UNI, and CTransPath as the best feature extractors in terms of achieving the lowest normalised differential \gls{auroc} scores averaged across all tasks, and found this to be the case for all three downstream models. 
Yet, it remains to be seen \emph{which} downstream model  is superior. 
To answer this question, we employ the technique from \cref{sec:feature_extractor_comparison}, but instead of keeping fixed the downstream model and determining the best feature extractor, we choose a feature extractor and vary the downstream model.
As shown in \cref{fig:norm_diff_auroc_downstream_model}, AttMIL performs best, closely followed by the transformer, and finally mean pooling, but we note the differences are small and exhibit high variance.

\subsection{Slide magnification}
The results presented so far utilise a slide magnification of $1.14$ \gls{mpp} (approximately $9\times$ magnification), which has been used in a number of prior works~\cite{kather2019deep,wagner2023transformerbased,ghaffari2022benchmarking,niehues2023generalizable,elnahhas2024regressionbased,yang2023devil,loeffler2023direct,elnahhas2023whole}
This is a popular choice in the literature, but by no means the only one; in fact, many studies use higher magnifications, such as $20\times$ ($0.5$ \acs{mpp})~\cite{coudray2018classification,lu2021dataefficient,liu2023identification}.

To assess the impact of slide magnification on downstream performance, we perform additional experiments at $0.5$ \gls{mpp} using the same setup as before.
In this additional set of experiments, we consider all feature extractors and tasks as before, but restrict our analysis to the AttMIL~\cite{ilse2018attention} aggregation model (which performed best in the previous section), and only consider two preprocessing setups: no augmentations and patchwise stain normalisation.

\begin{figure}[t]
  \centering
  \includegraphics[width=\linewidth]{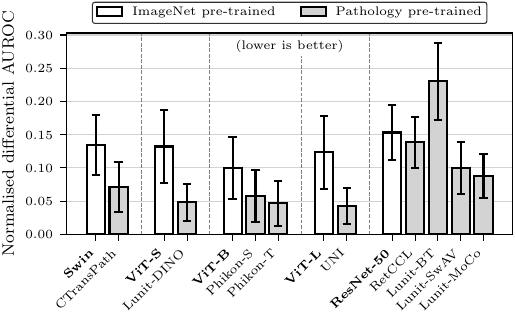}
  \caption{%
    Task-averaged relative performance comparison of the evaluated feature extractors at \emph{high magnification} ($20\times$), using the AttMIL aggregation model.
    This is analogous to \cref{fig:feature_extractor_performance_comparison}, which shows the relative performance at low magnification ($\approx 9\times$).
  }
  \label{fig:feature_extractor_performance_comparison_high_mag}
\end{figure}

\subsubsection{Relative performance at high magnification}
In \cref{fig:feature_extractor_performance_comparison_high_mag}, we compare the relative performance of the feature extractors at high magnification using the normalised differential \gls{auroc} score (\cf \cref{sec:feature_extractor_comparison}).
The top three feature extractors at high magnification are UNI~\cite{chen2024uni}, Phikon-T~\cite{filiot2023scaling}, and Lunit-DINO~\cite{kang2023benchmarking}, in that order.
Comparing to the low magnification results in \cref{fig:feature_extractor_performance_comparison}, we find that both Phikon~\cite{filiot2023scaling} models rank better at high magnification, with the teacher model now ranking second overall.
However, CTransPath~\cite{wang2022transformer}, which ranked third at low magnification, now ranks fifth.
Nonetheless, UNI~\cite{chen2024uni} and Lunit-DINO~\cite{kang2023benchmarking} remain among the top three feature extractors, though we note that there is no clear winner, with the relative performance differences among the top three being small, similar to the results at low magnification.

\begin{figure}
  \centering
  \includegraphics[width=\linewidth]{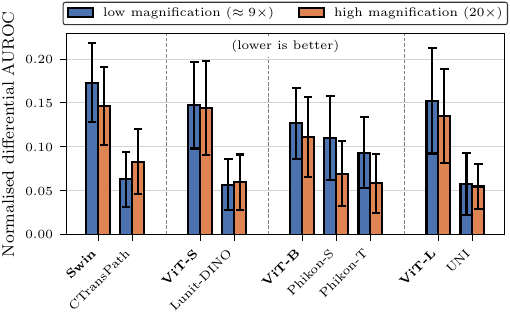}
  \caption{%
    Task-averaged relative performance comparison of selected feature extractors at \textcolor{figblue}{low} and \textcolor{figorange}{high} magnification, using no augmentations and downstream AttMIL~\cite{ilse2018attention} aggregation.
    This direct comparison of these 18 configurations (the product of nine feature extractors and two magnification levels) shows how the top-performing feature extractors fare at different magnifications, relative to each other.
  }
  \label{fig:feature_extractor_performance_comparison_best_low_and_high_mag}
\end{figure}

\subsubsection{Determining the best overall feature extractor and magnification}
The normalised differential \gls{auroc} score is a \emph{relative} performance measure.
While the rankings in \cref{fig:feature_extractor_performance_comparison,fig:feature_extractor_performance_comparison_high_mag} show which feature extractors are best at either high or low magnification, they cannot show which combination of feature extractor and magnification is best overall.
For this, we compute another set of normalised differential \gls{auroc} scores, but consider each feature extractor twice: once at low magnification and once at high magnification.
We do not include the ResNet-based feature extractors in this analysis, as they performed worst at both magnifications, to reduce the cost of computing the normalised differential \gls{auroc} scores -- the nine transformer-based feature extractors already require $5^{18}$ ($\approx$ 3 trillion) comparison operations.

The results, shown in \cref{fig:feature_extractor_performance_comparison_best_low_and_high_mag}, indicate that UNI~\cite{chen2024uni} and Lunit-DINO~\cite{kang2023benchmarking} are the best feature extractors at both magnifications, with the relative performances being very similar for both feature extractors across both magnifications.
UNI performs slightly better at high magnification, and Lunit-DINO slightly better at low magnification, with the former being the best overall.
However, the differences in performance among the top two feature extractors, being small in magnitude, are overshadowed by the variance of the scores.
As such, we conclude that there is no clear winner between the two magnifications; the choice of feature extractor (\ie preferring UNI/Lunit-DINO over the others) is more important than the magnification level.

Interestingly, the choice of magnification becomes important for the other feature extractors:
the Phikon models~\cite{filiot2023scaling} perform significantly better at high magnification, with the teacher model's high magnification version almost matching UNI and Lunit-DINO in performance.
On the other hand, CTransPath~\cite{wang2022transformer} performs worse at high magnification.

\begin{figure}[t]
  \centering
  \includegraphics[width=\linewidth]{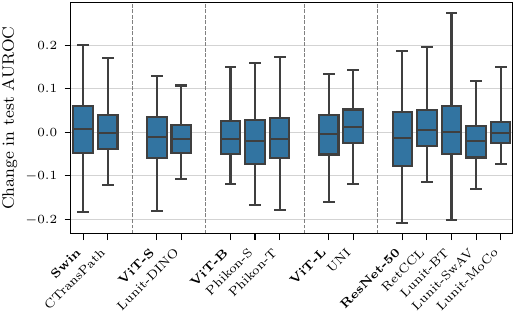}
  \caption{%
    Expected difference in test \gls{auroc} when performing patchwise stain normalisation \vs not at high magnification ($20\times$), using the AttMIL aggregation model.
    Positive values indicate a benefit from stain normalisation.
  }
  \label{fig:augmentations_performance_effect_high_mag}
  \vspace{-10pt}
\end{figure}

\subsubsection{Stain normalisation at high magnification}
In \cref{fig:augmentations_performance_effect_high_mag}, we observe no consistent benefit in employing patchwise stain normalisation at high magnification, especially for the top feature extractors.
This is in line with the results at low magnification, where we found no consistent performance differences between models trained with and without stain normalisation (\cf \cref{fig:augmentations_performance_effect}a).
We present extended results and analyses of stain normalisation at low \vs high magnification in \cref{sec:app:extended_results_high_mag}.

\section{Discussion}
\label{sec:discussion}
We dedicate this section to answer key questions that may arise among computational pathology researchers about employing \gls{ssl} feature extractors for slide-level prediction.

\vspace{5pt}\noindent\emph{What is the best pathology feature extractor?}
We recommend Lunit-DINO, UNI, and CTransPath, since they consistently achieve the best task-averaged downstream performance (\cref{fig:feature_extractor_performance_comparison}), independent of the employed augmentations and downstream aggregation model. 
In general, we find pathology-specific extractors outperform their ImageNet baselines, adding to the body of evidence that \gls{ssl} models pretrained on pathology data produce more useful features~\cite{shao2023generalizability,kang2023benchmarking,campanella2023computational,chen2021selfsupervisedhistopathology,sikaroudi2023generalization}. 

\vspace{5pt}\noindent\emph{At what level of magnification should features be extracted?}
The top feature extractors, UNI and Lunit-DINO, perform similarly at low and high magnification (\cref{fig:feature_extractor_performance_comparison_best_low_and_high_mag}), thus it is sufficient to extract features at low magnification ($\approx 9\times$) to reduce the number of patches and computational cost.

\vspace{5pt}\noindent\emph{Which aggregation model should we use?}
For Lunit-DINO, UNI, and CTransPath, we notice slight benefits in employing the AttMIL aggregation model downstream, though the differences are small and exhibit high variance (\cref{fig:norm_diff_auroc_downstream_model}).
The primary factor remains the choice feature extractor.

\vspace{5pt}\noindent\emph{Should we perform stain normalisation and augmentations?}
Our data does not support the necessity of either: they do not markedly improve performance, whilst incurring significant overhead.
When nonetheless employing stain normalisation, the patchwise approach should be preferred due to its lower computational cost. 
Image augmentations should always be avoided because in addition to the preprocessing cost, they considerably increase training time (\cref{sec:app:caching}), and the top extractors resist pathology-relevant augmentations (\cref{sec:latent_augmentations}).

\section{Conclusion}
In this work, we perform the most comprehensive robustness evaluation of publicly available pathology feature extractors for slide-level prediction to date, spanning 14 feature extractors, three aggregation models, stain normalisation, and numerous image augmentations, on nine downstream weakly supervised learning tasks with external validation cohorts. 
Among these factors, we find the choice of feature extractor is most consequential for downstream performance, and observe no benefit in employing stain normalisation or image augmentations. 
Further research is needed to understand the impact on computational pathology tasks other than \gls{wsi} classification, such as tumour segmentation.

Our latent space analysis reveals a remarkable robustness to stain variations and image augmentations in the top-performing feature extractors, Lunit-DINO~\cite{kang2023benchmarking}, UNI~\cite{chen2024uni}, and CTransPath~\cite{wang2022transformer}, which employ domain-specific knowledge in their \gls{ssl} training regimes.
This underlines the importance for future research into the development of pathology feature extractors and foundation models to not only scale size and diversity of pretraining datasets, but also to tailor \gls{ssl} methods to the pathology domain, in order to effectively leverage this data. 

\section*{Acknowledgements}
GW is supported by Lothian NHS. This project received funding from the European Union's Horizon 2020 research and innovation programme under Grant Agreement No. 101017453 as part of the KATY project. This work is supported in part by the Industrial Centre for AI Research in Digital Diagnostics (iCAIRD) which is funded by Innovate UK on behalf of UK Research and Innovation (UKRI) (project number 104690).
JNK is supported by the German Cancer Aid (DECADE, 70115166), the German Federal Ministry of Education and Research (PEARL, 01KD2104C; CAMINO, 01EO2101; SWAG, 01KD2215A; TRANSFORM LIVER, 031L0312A; TANGERINE, 01KT2302 through ERA-NET Transcan), the German Academic Exchange Service (SECAI, 57616814), the German Federal Joint Committee (TransplantKI, 01VSF21048) the European Union’s Horizon Europe and innovation programme (ODELIA, 101057091; GENIAL, 101096312), the European Research Council (ERC; NADIR, 101114631) and the National Institute for Health and Care Research (NIHR, NIHR203331) Leeds Biomedical Research Centre. The views expressed are those of the author(s) and not necessarily those of the NHS, the NIHR or the Department of Health and Social Care. This work was funded by the European Union. Views and opinions expressed are however those of the author(s) only and do not necessarily reflect those of the European Union. Neither the European Union nor the granting authority can be held responsible for them.

\printbibliography

\begin{IEEEbiographynophoto}{Georg W\"{o}lflein}
  received the MSci degree in Computer Science from the University of St Andrews, UK, in 2021.
  He is currently working towards a PhD at the same institution, supervised by Dr.~Ognjen Arandjelovi\'{c}, focusing on deep learning in computational pathology.
\end{IEEEbiographynophoto}
\begin{IEEEbiographynophoto}{Dyke Ferber}
  is working as a resident doctor at National Center for Tumor Diseases Heidelberg and a member of Prof.\ Jakob N.\ Kather's group at EKFZ Dresden. His research focuses on leveraging and improving image and text based AI models on real world data for oncology.
\end{IEEEbiographynophoto}
\begin{IEEEbiographynophoto}{Asier Rabasco Meneghetti}
  is a postdoctoral researcher at EKFZ in TUD within Prof.\ Jakob N.\ Kather's group. His research focuses in using machine and deep learning in radiological images for biomarker development alongside survival and statistical modelling.
\end{IEEEbiographynophoto}
\begin{IEEEbiographynophoto}{Omar S.\ M.\ El Nahhas}
   is an engineer who joined the Kather Lab for Clinical AI in 2022 led by Prof.\ Dr.\ Jakob N.\ Kather. He is currently pursuing a PhD in computer science, specializing in computational pathology. His research focuses on improving the performance and interpretability of deep learning in clinical AI. He is co-supervised by Prof.\ Dr.\ Ivo Sbalzarini from the faculty of Computer Science at TU Dresden.
\end{IEEEbiographynophoto}
\begin{IEEEbiographynophoto}{Professor Daniel Truhn}
  is senior physician in radiology at the University Hospital of Aachen in Germany. His background is in medicine and physics and he is leading a group of engineers, physicists and medical doctors who collaborate on the application of AI models to clinical practice.
\end{IEEEbiographynophoto}
\begin{IEEEbiographynophoto}{Zunamys I.\ Carrero}
  is a scientist with a PhD in Biochemistry, specializing in Molecular Oncology, and a BSc in Industrial Biotechnology. Her expertise in Precision Oncology includes developing patient-derived pre-clinical models and pharmacotyping, as well as working closely with clinicians to enhance the interpretation of drug-response data and tailor experimental designs. Since 2022, she has been a member of Prof.\ Dr.\ Jakob N.\ Kather’s lab leadership team, managing collaboration projects, scientific staff, and overall team strategy.
\end{IEEEbiographynophoto}
\begin{IEEEbiographynophoto}{Professor David J.\ Harrison}
   is a medical histopathologist interested in responses of tissue and cells to injury, and the application of computer vision and machine learning to re-invent what it means to extract useful information from tissue samples, rather than simply augment what a pathologist already does.
\end{IEEEbiographynophoto}
\begin{IEEEbiographynophoto}{Ognjen Arandjelovi\'{c}}
  graduated top of his class from the University of Oxford (M.Eng.). He was awarded his Ph.D.\ by the University of Cambridge, where he stayed thereafter as a Fellow of Trinity College. Presently he is a Reader at the University of St Andrews in Scotland. To date, Ognjen's work includes over 200 peer-reviewed publications, and is characterized by a highly polymathic nature. Ognjen is a Fellow of the Cambridge Overseas Trust, the winner of numerous awards, Associate Editor-in-Chief of \emph{Pattern Recognition}, and Associate Editor of \emph{Information} and \emph{Cancers}.
\end{IEEEbiographynophoto}
\begin{IEEEbiographynophoto}{Professor Jakob Nikolas Kather}
  is a senior physician in medical oncology at the University Hospital Dresden and holds appointments in medicine and computer science at TU Dresden, with an additional affiliation with the National Center for Tumor Diseases in Heidelberg. His research focuses on using artificial intelligence in precision oncology, applying deep learning to various clinical data types. His team combines medical and tech expertise, training medical researchers in programming and computer scientists in oncology.
\end{IEEEbiographynophoto}

\vfill

\clearpage
{\appendices
\begin{refsection}
\crefalias{section}{appendix}
\crefalias{subsection}{appendix}

\section{Downstream tasks and their clinical relevance}
\label{app:downstream_tasks}
\subsection{Targets}
We extensively evaluated the models on nine downstream tasks, summarised in \cref{tbl:app:tasks}.
All of the targets were treated as binary variables, except for breast cancer subtype, which is a five-way classification target, determined by immunohistochemistry: Luminal A (HR+/HER2-/low Ki-67), Luminal B (HR+/HER2+/high Ki-67), HER2 overexpressed (HR-), Basal (which is a subgroup of triple-negative breast cancer), or Normal breast-like (a subtype for which the clinical and molecular characteristics remain largely undefined throughout the existing scientific literature)~\cite{sorlie2001gene}. 
This molecular subtyping of early-stage invasive breast cancer has become an essential procedure in clinical management due to its implications in treatment recommendations and providing valuable prognostic insights for a patient's survival~\cite{cardoso2019early,goldhirsch2013personalizing}. 
In addition, our investigation also included analysis for prevalent mutations in CDH1 and TP53 as well as PIK3CA, the latter of which opens new possibilities for targeted therapies in advanced disease stages~\cite{andre2019alpelisib}. 
\Gls{msi} status is a key marker in colorectal cancer owing to its profound implications in shaping a patient's prognosis and responsiveness to immunotherapies~\cite{chalabi2022lba7,fda2017pembrolizumab}. 
It is driven by either spontaneous or germline (hence hereditary) mutations in DNA-repair related genes~\cite{buecher2013role} and leads to phenotypic changes in the tumour tissue~\cite{smyrk2001tumor}. 
Therefore, the performance of various AI models is commonly evaluated based on their ability to predict \gls{msi} from routine histopathology~\cite{kather2019deep}, often performed in conjunction with other prevalent genetic markers such as KRAS and BRAF: these are key-driver mutations in colorectal cancer, that shape a patient's survival chances and hold strong influence over the selection of targeted therapies best suited for each individual patient~\cite{roth2010prognostic,dienstmann2017prediction}. 
Given the high clinical relevance and availability of robust ground truth data, we have strategically selected these particular tasks for our analysis.

\begin{table*}[b]
  \caption{Overview of the evaluated downstream tasks. Dataset sizes are shown in parentheses. The \breasticon~targets are related to breast cancer, while the \colonicon~targets are related to colorectal cancer.}
  \label{tbl:app:tasks}
  \centering
  \begin{tabular}{p{3cm}|p{4.2cm}p{4.2cm}}
      \toprule
      \textbf{Target} & \textbf{Training and validation} & \textbf{Test dataset} \\
      \midrule
      \breasticon-Subtype & \multirow{4}{*}{\shortstack[l]{\acs{tcga}-BRCA~\cite{weinstein2013cancer}\\(833 train, 208 val samples)}} & \multirow{4}{*}{\shortstack[l]{CPTAC-BRCA~\cite{krug2020proteogenomic}\\(120 samples)}} \\
      \breasticon-CDH1 mutation & & \\
      \breasticon-TP53 mutation & & \\
      \breasticon-PIK3CA mutation & & \\
      \midrule
      \breasticon-LN status & \parbox{4.2cm}{\raggedright CAMELYON17~\cite{bandi2018detection}\\(centre-wise cross-validation; 320 train, 80 val samples)} & \parbox{4.2cm}{\raggedright CAMELYON17~\cite{bandi2018detection}\\(centre-wise cross-validation; 100 samples)} \\   
      \midrule
      \colonicon-\acs{msi} status & \multirow{4}{*}{\shortstack[l]{\acs{tcga}-CRC~\cite{weinstein2013cancer}\\(558 samples)}} & \multirow{4}{*}{\shortstack[l]{CPTAC-COAD~\cite{vasaikar2019proteogenomic}\\(110 samples)}} \\
      \colonicon-RAS mutation & & \\
      \colonicon-RAF mutation & & \\
      \colonicon-MAD4 mutation & & \\
      \bottomrule
  \end{tabular}
\end{table*}

\subsection{Data}
Here, we provide additional details about where we obtained data for the downstream tasks, further to what is mentioned in \cref{sec:downstream}.

We predict the \breasticon-LN status using the CAMELYON17 dataset~\cite{bandi2018detection}, which contains data from five centres.
For this dataset, we perform centre-wise cross-validation, where we use one of the centres for testing and the others for training (each of the five random seeds uses a different centre for testing).
The training and validation sets are an 80\%/20\% split of the other four centres.
We treat \breasticon-LN status as a binary classification task, where the positive class corresponds to the presence of metastatic cancer cells in the lymph nodes.
Each slide in the dataset is of a lymph node tissue section, and we treat each slide as a single sample, \ie a separate patient.
This is slightly different to the original CAMELYON17 challenge~\cite{bandi2018detection}, where groups of five slides were arranged into ``virtual patients'' (though the slides themselves may be from different actual patients), and the task was to predict a virtual patient-level label based on a specific rule for aggregating the slide-level predictions.
We do not use the virtual patient labels, but instead use the slide-level labels provided in the dataset.

For all other targets, we use either \acs{tcga}-BRCA~\cite{weinstein2013cancer} or \acs{tcga}-CRC~\cite{weinstein2013cancer} for training, and respectively either CPTAC-BRCA~\cite{krug2020proteogenomic} or CPTAC-COAD~\cite{vasaikar2019proteogenomic} for testing.
We obtain the patient-level labels from the respective studies via \url{cbioportal.org}.
The only exception is \colonicon-\acs{msi} status which is not available for \acs{tcga}-CRC in \url{cbioportal.org}, but is provided in the supplementary material of Liu \etal~\cite{liu2018comparative}.

\subsection{Human subject data}
The aforementioned datasets (\acs{tcga}-BRCA~\cite{weinstein2013cancer}, \acs{tcga}-CRC~\cite{weinstein2013cancer}, CPTAC-BRCA~\cite{krug2020proteogenomic}, CPTAC-COAD~\cite{vasaikar2019proteogenomic}, and CAMELYON17~\cite{bandi2018detection}) contain data from human subjects.
These datasets are publicly available and have been de-identified.

\section{Feature extractors}
\label{sec:app:feature_extractors}
In \cref{tbl:app:feature_extractors_overview}, we provide an overview of the \gls{ssl} feature extractors evaluated in this study.
We use the weights from the respective authors' GitHub repositories. 
The feature extractor called Lunit-DINO in our paper corresponds to Kang \etal's DINO$_{p=16}$ model~\cite{kang2023benchmarking}. 
For the Phikon extractor~\cite{filiot2023scaling}, we employ both the `student' (Phikon-S) and `teacher' (Phikon-T) models.
\begin{table*}
  \caption{%
    Overview of \gls{ssl} feature extractors evaluated in this study, their architecture, \gls{ssl} method, pretraining dataset, and embedding size. 
    As baselines, we additionally compare against the respective ImageNet pretrained backbones: Swin Transformer~\cite{liu2021swin}, Vit-B~\cite{kolesnikov2021image}, ViT-S~\cite{kolesnikov2021image,steiner2021how} and ResNet-50~\cite{he2015deep}.
  }
  \label{tbl:app:feature_extractors_overview}
  \resizebox{\linewidth}{!}{%
    \begin{tabular}{llp{3.5cm}p{5cm}c}
      \toprule
      \textbf{Name} & \textbf{Architecture} & \textbf{\gls{ssl} method} & \textbf{\gls{ssl} dataset, magnification} & \textbf{Embedding size ($d_x$)} \\
      \midrule
      CTransPath~\cite{wang2022transformer,wang2021transpath} & Swin Transformer~\cite{liu2021swin} & \acl{srcl}~\cite{wang2022transformer} based on MoCo v3~\cite{chen2021empirical} & \acs{tcga}~\cite{weinstein2013cancer} and PAIP~\cite{kim2021paip} ($20\times$) & 768 \\
      Phikon-S~\cite{filiot2023scaling} & ViT-B~\cite{kolesnikov2021image} & iBOT~\cite{zhou2022image} & \acs{tcga}~\cite{weinstein2013cancer} ($20\times$) & 768 \\
      Phikon-T~\cite{filiot2023scaling} & ViT-B~\cite{kolesnikov2021image} & iBOT~\cite{zhou2022image} & \acs{tcga}~\cite{weinstein2013cancer} ($20\times$) & 768 \\
      UNI~\cite{chen2024uni} & ViT-L~\cite{kolesnikov2021image} & DINOv2~\cite{oquab2023dinov2} & non-public MASS-100K~\cite{chen2024uni} ($20\times$) & 1024 \\
      Lunit-DINO~\cite{kang2023benchmarking} & ViT-S~\cite{kolesnikov2021image} & DINO~\cite{caron2021emerging} & \acs{tcga}~\cite{weinstein2013cancer} and non-public TULIP~\cite{kang2023benchmarking} ($20\times,40\times$) & 384 \\
      RetCCL~\cite{wang2023retccl} & ResNet-50~\cite{he2015deep} & \acl{ccl}~\cite{wang2023retccl} based on MoCo~\cite{he2020momentum} & \acs{tcga}~\cite{weinstein2013cancer} and PAIP~\cite{kim2021paip} ($20\times$) & 2048 \\
      Lunit-BT~\cite{kang2023benchmarking} & ResNet-50~\cite{he2015deep} & Barlow Twins~\cite{zbontar2021barlow} & \acs{tcga}~\cite{weinstein2013cancer} and non-public TULIP~\cite{kang2023benchmarking} ($20\times,40\times$) & 2048 \\
      Lunit-SwAV~\cite{kang2023benchmarking} & ResNet-50~\cite{he2015deep} & SwAV~\cite{caron2020unsupervised} & \acs{tcga}~\cite{weinstein2013cancer} and non-public TULIP~\cite{kang2023benchmarking} ($20\times,40\times$) & 2048 \\
      Lunit-MoCo~\cite{kang2023benchmarking} & ResNet-50~\cite{he2015deep} & MoCo~v2~\cite{chen2020improved} & \acs{tcga}~\cite{weinstein2013cancer} and non-public TULIP~\cite{kang2023benchmarking} ($20\times,40\times$) & 2048 \\
      \bottomrule
    \end{tabular}
  }
\end{table*}

\subsection{Foundation models}
\label{sec:app:foundation_models}
This year, a number of foundation models have emerged for pathology that were trained on datasets of unprecedented size. 
Of these, we were only able to include UNI~\cite{chen2024uni} in our study, as the others' weights remain proprietary.
Nonetheless, we briefly review these pathology foundation models below.
UNI~\cite{chen2024uni} and RudolfV~\cite{dippel2024rudolfv} have been trained on datasets exceeding 100,000 slides, while Virchow~\cite{vorontsov2023virchow} utilises an even larger corpus of 1.5 million slides, all three employing the DINOv2 framework~\cite{oquab2023dinov2}.
Moreover, Campanella \etal~\cite{campanella2023computational} trained two foundation models on over 400,000 slides using DINO~\cite{caron2021emerging} and MAE~\cite{he2021masked}.
On the other hand, Azizi \etal~\cite{azizi2023robust} integrate both medical and non-medical images to train their foundation model, REMEDIS, using SimCLR/BiT~\cite{chen2020simple,chen2020big}.
Furthermore, Lu \etal~\cite{lu2024conch} made use of 1.17 million image-caption pairs to develop a vision-language foundation model named CONCH.
In stark contrast, the publicly available models (with the notable exception of UNI~\cite{chen2024uni}) employ orders of magnitude fewer \glspl{wsi}, as \gls{tcga} contains around 30,000 diagnostic and tissue slides in total~\cite{weinstein2013cancer}.

\section{Stain normalisation}
\label{sec:app:stain_normalisation}
In \cref{fig:app:latent_macenko_all}, we show the effect of stain normalisation on the latent space of all 14 feature extractors, extending \cref{fig:latent_macenko} from the main text which showed just two. 
\begin{figure*}
    \includegraphics[width=\textwidth]{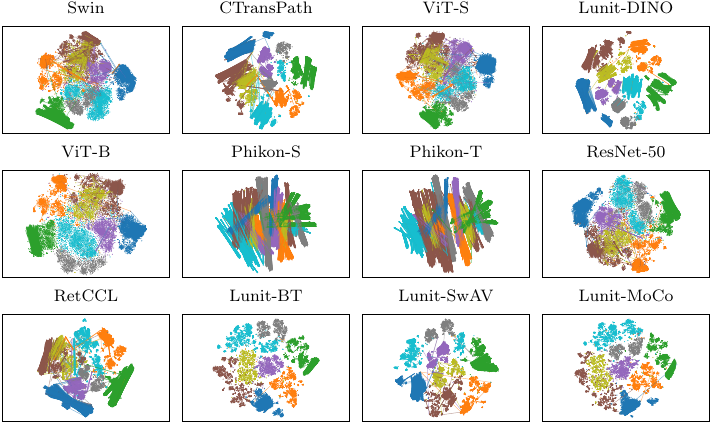}
    \caption{%
        Latent space visualisations (t-SNE~\cite{vandermaaten2008visualizing}), showing the effect of stain normalisation~\cite{macenko2009method}.
        This figure extends \cref{fig:latent_macenko}, which depicts only two feature extractors, Lunit-DINO~\cite{kang2023benchmarking} and its ViT-S~\cite{kolesnikov2021image,steiner2021how} ImageNet baseline; here, we show all evaluated feature extractors.
        Colours are as in \cref{fig:latent_macenko}.
    }
    \label{fig:app:latent_macenko_all}
\end{figure*}

\subsection{Patchwise versus slidewise stain normalisation}
In \cref{sec:stain_normalisation_downstream}, we state that there is no consistent improvement obtained by employing stain normalisation, regardless of whether it is performed on a per-patch or per-slide basis.
Further to the results in \cref{fig:augmentations_performance_effect}a showing only slidewise stain normalisation, we perform an ablation study where we normalise each patch individually. 
We provide an analogous boxplot for both types of stain normalisation in \cref{fig:stainnorm_ablation}, which shows that the conclusion holds for both types of stain normalisation.
\begin{figure}
    \centering
    \includegraphics[width=\linewidth]{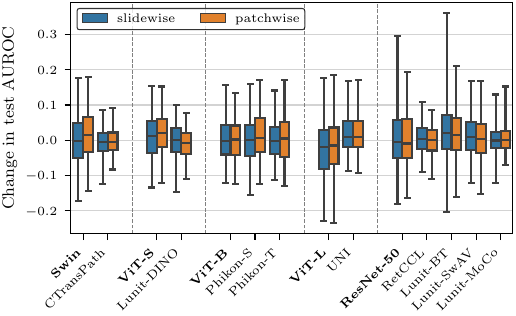}
    \caption{%
        Improvement obtained by employing slidewise (\textcolor{figblue}{blue}, boxes are as in \cref{fig:augmentations_performance_effect}a) or patchwise (\textcolor{figorange}{orange}) stain normalisation compared to no normalisation.
        There is no clear benefit or detriment in applying either type of stain normalisation (all confidence intervals cross zero).
        While this figure reports results only for the downstream AttMIL model, but we observe a similar phenomenon for the other two aggregation models.
    }
    \label{fig:stainnorm_ablation}
\end{figure}


\section{Augmentations}
\label{sec:app:augmentations}
\begin{figure*}
    \includegraphics[width=\textwidth]{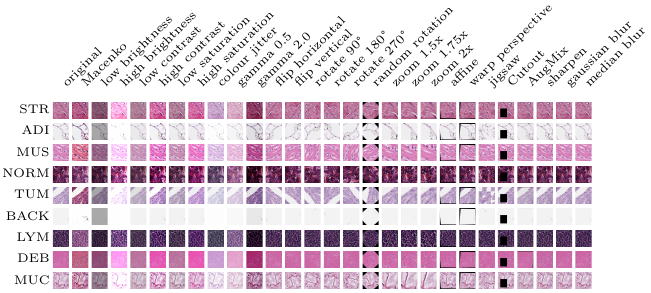}
    \caption{%
        Examples of original and augmented patches (columns) from the NCT-CRC-HE-100K dataset~\cite{kather2018100000,kather2019predicting}.
        Each row corresponds to a representative patch from a different patch class.
    }
    \label{fig:app:augmentations_overview}
\end{figure*}
Including patchwise stain normalisation, we study 27 image augmentations in this work. We provide representative examples of these augmentations in \cref{fig:app:augmentations_overview}, and describe them below:
\begin{itemize}
    \item \textbf{macenko}: Macenko stain normalisation~\cite{macenko2009method} (patchwise)
    \item \textbf{rotate \{90°, 180°, 270°\}}: rotate by the specified angle
    \item \textbf{random rotation}: rotate by an angle $\beta$ sampled uniformly such that $(\beta \mod 90) \in [10, 80]$, \ie forcing an off-axis rotation
    \item \textbf{flip \{horizontal, vertical\}}: flip along the specified axis
    \item \textbf{zoom \{1.5$\times$, 1.75$\times$, 2$\times$\}}: enlarge the patch by the specified factor and crop the centre
    \item \textbf{affine}: random affine transformation with a maximum rotation of 10°, maximum translation of 20\% of the patch size, maximum scaling of 20\%, and maximum shear of 10°
    \item \textbf{warp perspective}: random perspective transformation with a maximum distortion of 0.2
    \item \textbf{jigsaw}: cut the patch into a $4 \times 4$ grid and randomly permute the tiles
    \item \textbf{Cutout}: randomly erase a rectangle that covers between 2\% and 25\% of the total area~\cite{devries2017improved}
    \item \textbf{AugMix}: see Hendrycks \etal~\cite{hendrycks2020augmix}
    \item \textbf{\{low, high\} brightness}: reduce the brightness by a factor of 0.7 or increase it by a factor of 1.5
    \item \textbf{\{low, high\} contrast}: reduce the contrast by a factor of 0.7 or increase it by a factor of 1.5
    \item \textbf{\{low, high\} saturation}: reduce the saturation by a factor of 0.7 or increase it by a factor of 1.5
    \item \textbf{colour jitter}: randomly adjust the brightness, contrast, saturation, and hue by maximum factors of 0.4, 0.4, 0.4, and 0.1, respectively
    \item \textbf{gamma \{0.5, 2.0\}}: apply a gamma correction with the specified exponent
    \item \textbf{sharpen}: sharpen the image by a factor of 5
    \item \textbf{Gaussian blur}: apply a Gaussian blur with a kernel size of 5 and a standard deviation of 2.0
    \item \textbf{median blur}: apply a median blur with a kernel size of 5
\end{itemize}

\subsection{Augmentation groups}
In \cref{sec:downstream}, we study the effect of various \emph{groups} of augmentations on downstream performance. These groups are defined as follows:
\begin{itemize}
    \item \textbf{none}: no augmentations, \ie the original patches are used
    \item \textbf{Macenko (patchwise)}: Macenko stain normalisation~\cite{macenko2009method} is applied on a per-patch basis
    \item \textbf{Macenko (slidewise)}: Macenko stain normalisation~\cite{macenko2009method} is applied on a per-slide basis
    \item \textbf{rotation/flipping}: each patch is randomly rotated by a right angle or flipped along the horizontal or vertical axis, with equal probability
    \item \textbf{all}: any of the 27 augmentations, or no augmentation, is applied to each patch with equal probability
\end{itemize}
We apply no augmentations to the test set (except when applying slidewise or patchwise stain normalisation, in which case we normalise the test set in the same way as the training set).

\section{Extended data tables and figures}
\label{sec:app:extended_data_and_figures}
In much of our discussion in \cref{sec:downstream,sec:discussion}, we focus on particular augmentations, models, feature extractors, and magnifications.
Here, we produce extended versions of figures and tables from the main text providing more results for different choices of the above.

\Cref{fig:feature_extractor_performance_comparison_extended,fig:augmentations_performance_effect_extended} summarises the main results for the remaining two downstream aggregation models (whereas \cref{fig:feature_extractor_performance_comparison,fig:augmentations_performance_effect} from the text focused on AttMIL~\cite{ilse2018attention}): the two-layer transformer as employed by Wagner \etal~\cite{wagner2023transformerbased}, and the mean average pooling baseline.
\begin{figure*}
  \centering
  \begin{subfigure}[t]{.49\linewidth}
    \centering
    \includegraphics[width=\linewidth]{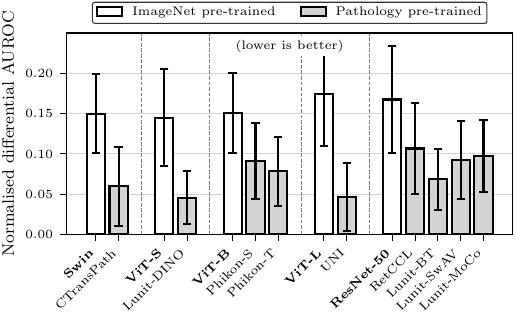}
    \caption{Two-layer transformer~\cite{wagner2023transformerbased}}
  \end{subfigure}
  \begin{subfigure}[t]{.49\linewidth}
    \centering
    \includegraphics[width=\linewidth]{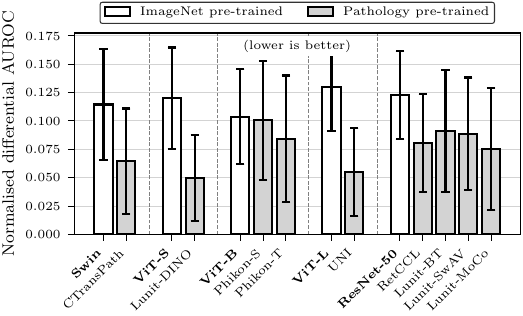}
    \caption{Mean average pooling}
  \end{subfigure}
  \caption{%
      Extended version of \cref{fig:feature_extractor_performance_comparison} showing the performance of the remaining two downstream models: a two-layer transformer~\cite{wagner2023transformerbased} (left) and mean average pooling (right).
  }
  \label{fig:feature_extractor_performance_comparison_extended}
\end{figure*}
\begin{figure*}
  \centering
  \includegraphics[width=\linewidth]{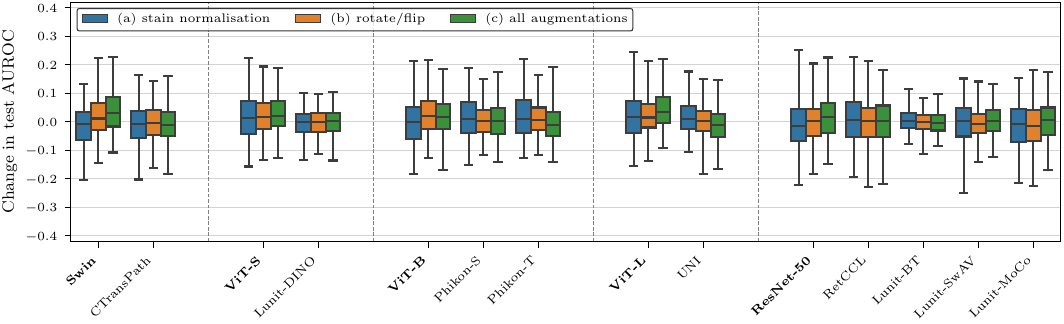}
  \includegraphics[width=\linewidth]{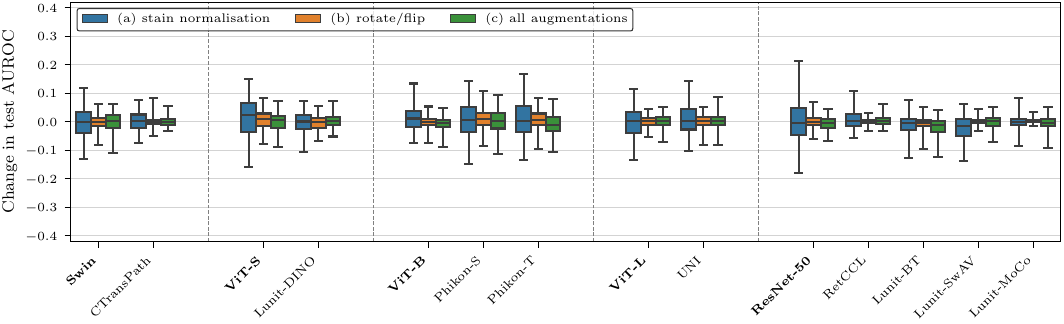}
  \caption{%
      Extended version of \cref{fig:augmentations_performance_effect} showing the impact of augmentations on downstream performance for the remaining two downstream models: a two-layer transformer~\cite{wagner2023transformerbased} (top) and mean average pooling (bottom).
  }
  \label{fig:augmentations_performance_effect_extended}
\end{figure*}

\subsection{Normalised differential \gls{auroc} scores}
In \cref{tbl:extended_data_norm_diff_auroc_none,tbl:extended_data_norm_diff_auroc_Macenko_slidewise,tbl:extended_data_norm_diff_auroc_Macenko_patchwise,tbl:extended_data_norm_diff_auroc_simple_rotate,tbl:extended_data_norm_diff_auroc_all}, we present the normalised differential \gls{auroc} scores for all tasks, feature extractors, and downstream models, and augmentation groups (one table per augmentation group).
This extends \cref{tbl:norm_differential_auroc_attmil_noaug} from the main text, which only shows the results for the AttMIL~\cite{ilse2018attention} aggregation model without augmentations (corresponding to the first 14 rows in \cref{tbl:extended_data_norm_diff_auroc_none}).
We observe that Lunit-DINO~\cite{kang2023benchmarking} and CTransPath~\cite{wang2022transformer} consistently achieve the best task-averaged results, independent of the choice of downstream aggregation model and augmentation group.

\subsection{Absolute \gls{auroc} scores}
While the normalised differential \gls{auroc} score provides a relative performance measure to facilitate a fair comparison between feature extractors, we also provide the seed-averaged absolute test \gls{auroc} scores for all tasks, feature extractors, and downstream models, and augmentation groups in \cref{tbl:extended_data_norm_diff_auroc_none,tbl:extended_data_norm_diff_auroc_Macenko_slidewise,tbl:extended_data_norm_diff_auroc_Macenko_patchwise,tbl:extended_data_norm_diff_auroc_simple_rotate,tbl:extended_data_norm_diff_auroc_all} (one table per augmentation group).
Looking at these absolute scores, we find that the predicting the \breasticon-PIK3CA target is the most difficult task across the board for all feature extractors and downstream models, while the \breasticon-LN status and \colonicon-MSI status targets are the easiest.
However, we emphasise that the normalised differential \gls{auroc} score is the more meaningful metric for comparing feature extractors, since it is independent of the task difficulty and accounts for the variance across seeds (see \cref{sec:feature_extractor_comparison}).

\subsection{Results at high magnification}
\label{sec:app:extended_results_high_mag}
In \cref{tbl:extended_data_norm_diff_auroc_none_high,tbl:extended_data_norm_diff_auroc_Macenko_patchwise_high}, we provide the normalised differential \gls{auroc} scores for all tasks and feature extractors at high magnification ($0.5$ \acs{mpp}) which is analogous to \cref{tbl:extended_data_norm_diff_auroc_none,tbl:extended_data_norm_diff_auroc_Macenko_patchwise} at low magnification ($1.14$ \acs{mpp}).
Furthermore, we present the seed-averaged absolute test \gls{auroc} scores for high magnification in \cref{tbl:extended_data_mean_auroc_none,tbl:extended_data_mean_auroc_Macenko_patchwise}, which is analogous to \cref{tbl:extended_data_mean_auroc_none,tbl:extended_data_mean_auroc_Macenko_patchwise} at low magnification.

\begin{figure}
  \centering
  \includegraphics[width=\linewidth]{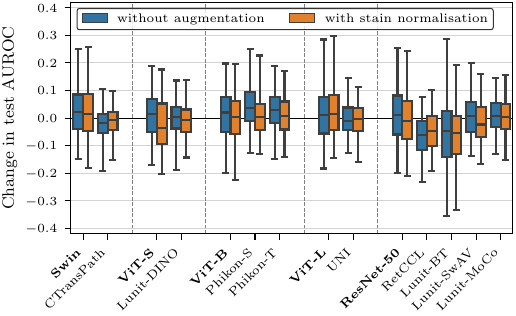}
  \caption{%
    Expected change in performance when training on low ($\approx 9\times)$ \vs high ($20\times$) magnification (positive values indicate better performance at high magnification).
    Shown in \textcolor{figblue}{blue} are the results without any augmentations, and in \textcolor{figorange}{orange} with patchwise stain normalisation.
    We observe no consistent performance benefit in training at high magnification, especially in the top-performing feature extractors, no matter the preprocessing (no augmentations or stain normalisation).
  }
  \label{fig:magnification_performance_effect}
\end{figure}

In \cref{fig:magnification_performance_effect}, we perform a bootstrapped comparison to quantify the change in performance when using each feature extractor at low compared to high magnification, akin to the analysis in \cref{fig:augmentations_performance_effect}.
The `no augmentation' results (\textcolor{figblue}{blue} in \cref{fig:magnification_performance_effect}) are consistent with the results in \cref{fig:feature_extractor_performance_comparison_best_low_and_high_mag}, showing no change in performance for the top feature extractors, UNI~\cite{chen2024uni} and Lunit-DINO~\cite{kang2023benchmarking}.
Even with stain normalisation (\textcolor{figorange}{orange} in \cref{fig:magnification_performance_effect}), the performance of the top feature extractors remains consistent across magnifications.
However, stain normalisation has a more pronounced effect on the Phikon models~\cite{filiot2023scaling} at high magnification:
without stain normalisation, Phikon-S and Phikon-T perform better at high than low magnification (an effect observed previously in \cref{fig:feature_extractor_performance_comparison_high_mag}, and now in \cref{fig:magnification_performance_effect}, \textcolor{figblue}{blue}), but with stain normalisation, magnification has a much weaker effect on performance (\textcolor{figorange}{orange}).
Apart from this, the magnification-related performance changes are largely unaffected by the choice of preprocessing (stain normalisation \vs not).

\section{Training and implementation details}
For downstream model training, we use the AdamW~\cite{loshchilov2017decoupled} optimiser with an initial learning rate of $10^{-3}$, weight decay of $10^{-2}$, and a batch size of one.
The learning rate is decayed using a cosine annealing learning schedule~\cite{loshchilov2016sgdr} over 30 epochs, but we halt training when the validation loss does not improve for ten epochs.

In \gls{mil} terminology, we refer to the \emph{patient} as the \emph{bag}, and the \emph{patches} as the \emph{instances}.
Note that some datasets have multiple \glspl{wsi} per patient; in these cases, we simply mix the patches from all \glspl{wsi} into a single bag.
An epoch represents a full pass over all patients in the training set. 
At every step, we sample a maximum of 8,192 patches per patient, though most patients have fewer patches.
We found it beneficial to employ a batch size of one: not only does this reduce GPU memory requirements, it also accelerates training. 
Indeed, we found that padding the bags to the maximum number of patches per patient (8,192) slows down training considerably, but with a batch size of one, we can use a variable number of patches per bag.
Nonetheless, we accumulate gradients over four steps before performing a weight update, which effectively increases the batch size to four.

\subsection{Downstream aggregation models}
We describe the three downstream aggregation models in more detail below.
In essence, these are different parametrisations of the $g_\theta$ function in \cref{eq:mil_aug} that aggregate the patch embeddings into a single slide-level embedding.
All three models first pass the patch embeddings through a linear layer with 512 output units and ReLU activation, \ie
\begin{equation}
    g_\theta(\{x_i\}_{i=1}^n) = \bar{g_\theta}(\{\max(0,\bar{W_\theta} x_i + \bar{b_\theta})\}_{i=1}^n)
\end{equation}
with learnable parameters $\bar{W_\theta} \in \mathbb{R}^{512 \times d_x}$ and $\bar{b_\theta} \in \mathbb{R}^{512}$.
However, the three models differ in how they aggregate the resulting patch embeddings in the $\bar{g_\theta}$ function, which we describe below.

In any case, the classifier $h_\theta$ in \cref{eq:mil_aug} is a linear layer with softmax activation over the number of classes, to which we apply a cross-entropy loss.
Note that we employ dropout with a probability of 0.5 to the slide-level embedding before passing it to the linear layer.

\subsubsection{Mean average pooling}
As a baseline model, we compute the slide-level embedding as the mean of the patch embeddings, \ie
\begin{equation}
    \label{eq:mean_average_pooling}
    \bar{g_\theta}(\{x_i\}_{i=1}^n) = \frac{1}{n} \sum_{i=1}^n x_i.
\end{equation}

\subsubsection{AttMIL~\cite{ilse2018attention}}
This model takes a weighted average of the patch embeddings, where the weights are computed independently for each embedding.
More formally, the slide-level embedding is given by
\begin{equation}
    \label{eq:attmil}
    \bar{g_\theta}(\{x_i\}_{i=1}^n) = \sum_{i=1}^n \alpha_i x_i,
\end{equation}
where the attention\footnote{Ilse \etal~\cite{ilse2018attention}'s use of the term ``attention'' should not be confused with the scaled dot product attention in the transformer architecture~\cite{vaswani2017attention}. Here, the attention weight for a particular token is computed solely based on that token alone.} weights $\alpha_i\in\mathbb{R}$ are obtained via a two-layer network with 256 $\tanh$-activated hidden units that is applied to each patch embedding $x_i$ independently and then normalised across all patches using a softmax function, \ie
\begin{align}
    \label{eq:attmil_alpha}
    e_i &= W_2 \tanh(W_1 x_i + b_1) + b_2, \\
    \alpha_i &= \frac{\exp(e_i)}{\sum_{j=1}^n \exp(e_j)}.
\end{align}
Here, $W_1 \in \mathbb{R}^{256 \times 512}$, $b_1 \in \mathbb{R}^{256}$, $W_2 \in \mathbb{R}^{1 \times 256}$, and $b_2 \in \mathbb{R}$ are learnable parameters (captured within the set of learnable parameters $\theta$).

\subsubsection{Two-layer transformer}
We also employ a two-layer transformer~\cite{vaswani2017attention}, closely aligned with the configuration presented by Wagner \etal~\cite{wagner2023transformerbased}.
This setup differs from the classical transformer architecture~\cite{vaswani2017attention} in that there is just one branch, \ie just the decoder (or encoder, depending on perspective).
Both layers have 512 hidden units and 8 attention heads, employ a dropout rate of 0.1, use GELU activation~\cite{hendrycks2016gelu} in the feedforward layers, and use layer normalisation~\cite{ba2016layer} before the attention layers.
We employ no masking.
The input tokens are the patch embeddings, and the output tokens are averaged like in \cref{eq:mean_average_pooling} to obtain the slide-level embedding.

\subsection{Overhead and caching}
\label{sec:app:caching}
\subsubsection{Feature extraction}
Prior to training, we extract features from all patches in the training and validation sets, and store them on disk.
We do this for each of the 14 feature extractors.
For the training sets, we additionally perform feature extraction for all 27 augmented versions of each patch, and store these on disk as well.
For both the training and test sets, we also extract features for the stain-normalised versions of the patches.
This way, we effectively have a cache of the $a_i \circ f$ function in \cref{eq:mil_aug} for all inputs (\ie patches), all augmentations $a_i$, and all feature extractors $f$.
During training, we only need to load the features from disk ($d_x$ floating point values per patch, \eg in the case of CTransPath $d_x=768$), as opposed to loading the patches directly ($224\times 224\times 3$ byte values) and having to perform augmentation and feature extraction on the fly (very expensive). 

\subsubsection{Training with augmentations}
Even though our training runs employed already extracted features, they took $30\times$ longer with all augmentations, or $5\times$ longer with just the rotation augmentations as compared to employing no augmentations. 
This approximately linear scaling in the number of augmentations $a$ is the result of slower data loading, as random reads are performed over $a$ times as many features compared to the no-augmentation case. 
We alleviated some of this bottleneck by implementing additional caches, but even this solution only bore fruit because we ran many experiments with similar dataset configurations.
Thus, we emphasise again that augmentations are too expensive to be viable in computational pathology pipelines, due their significant preprocessing \emph{and} training overhead which does not even yield a consistent improvement in downstream performance.

\subsubsection{Total training time}
We trained 9,450 models at low magnification ($1.14$ \acs{mpp}) across the cartesian product of:
\begin{itemize}
    \item 14 feature extractors,
    \item 5 augmentation groups,
    \item 3 downstream aggregation models,
    \item 9 downstream tasks, and
    \item 5 random seeds.
\end{itemize}
At high magnification ($0.5$ \acs{mpp}), we trained an additional 1,260 models; however, this was only for two augmentation groups (none and patchwise stain normalisation) and one downstream aggregation model (AttMIL~\cite{ilse2018attention}).
In total, we trained 10,710 models.

We trained these models on several NVIDIA Tesla V100 GPUs (one training run per GPU at a time), which cumulatively took 8,140 GPU hours (339.2 days) if it were to be run on a single GPU.
Of this, 5,990 GPU hours (249.6 days) were spent training models at low magnification ($1.14$ \acs{mpp}), and 2,150 GPU hours (89.6 days) were spent training models at high magnification ($0.5$ \acs{mpp}), though the latter was only for a subset of the models.
Note that these summary statistics do not include time spent on preprocessing and feature extraction.

\subsection{Normalised differential \gls{auroc} scores}
As explained in \cref{sec:feature_extractor_comparison}, the normalised differential \gls{auroc} score is a relative performance measure that accounts for the variance across seeds and the difficulty of the tasks.
For $S$ seeds and $F$ feature extractors, its computation requires $S^F$ iterations as per \cref{eq:norm_diff_auroc}.
Each iteration produces $F$ scores (one per feature extractor), which need to be aggregated across the $S^F$ iterations to obtain the final score for each feature extractor.
It is too expensive to keep all $FS^F$ intermediate scores in memory, so we instead compute the mean and standard deviation of the scores online using a batched streaming version of Welford's algorithm~\cite{welford1962note}.

For \cref{fig:feature_extractor_performance_comparison} with $S=5$ seeds and $F=14$ feature extractors, the computation requires $5^{14} \approx 6.1 \times 10^{9}$ iterations per downstream task, which is feasible to compute in a reasonable amount of time (about a day), on a single CPU with multiple threads.
However, this setup is infeasible for \cref{fig:feature_extractor_performance_comparison_best_low_and_high_mag}, which required the largest such computation with $S=5$ seeds and $F=18$ feature extractors ($5^{18} \approx 3.8 \times 10^{12}$ iterations per downstream task).
For this, we implemented a GPU kernel in Triton to compute the normalised differential \gls{auroc} scores in parallel, which took around four hours per downstream task on an NVIDIA A100 80GB GPU.

\begin{table*}[p]
  \caption{%
  Normalised differential \gls{auroc} scores for all tasks, feature extractors, downstream models, when employing \textbf{no augmentations} at \textbf{low magnification} ($1.14$ \acs{mpp}).
  }
  \label{tbl:extended_data_norm_diff_auroc_none}
  \resizebox{\textwidth}{!}{
\begin{tabular}{ll|cccc|c|cccc|c}
\toprule
 & Target & \breasticon-Subtype & \breasticon-CDH1 & \breasticon-TP53 & \breasticon-PIK3CA & \breasticon-LN status & \colonicon-MSI & \colonicon-KRAS & \colonicon-BRAF & \colonicon-SMAD4 & Average \\
Model & Feature extractor &  &  &  &  &  &  &  &  &  &  \\
\midrule
\multirow[t]{14}{*}{AttMIL} & Swin~\cite{liu2021swin} & $0.10 \pm 0.02$ & $0.17 \pm 0.02$ & $0.31 \pm 0.02$ & $0.07 \pm 0.04$ & $0.20 \pm 0.09$ & $0.18 \pm 0.04$ & $0.14 \pm 0.04$ & $0.14 \pm 0.06$ & $0.16 \pm 0.05$ & $0.163 \pm 0.047$ \\
 & CTransPath~\cite{wang2022transformer} & $0.03 \pm 0.02$ & $\mathbf{0.01 \pm 0.01}$ & $0.05 \pm 0.02$ & $0.05 \pm 0.03$ & $0.08 \pm 0.07$ & $0.09 \pm 0.03$ & $0.06 \pm 0.03$ & $0.06 \pm 0.03$ & $0.06 \pm 0.03$ & $0.053 \pm 0.034$ \\
 & ViT-S~\cite{kolesnikov2021image} & $0.16 \pm 0.02$ & $0.08 \pm 0.03$ & $0.17 \pm 0.05$ & $0.08 \pm 0.05$ & $0.22 \pm 0.09$ & $0.19 \pm 0.04$ & $0.06 \pm 0.03$ & $0.20 \pm 0.04$ & $0.08 \pm 0.08$ & $0.138 \pm 0.052$ \\
 & Lunit-DINO~\cite{kang2023benchmarking} & $0.10 \pm 0.02$ & $0.04 \pm 0.03$ & $0.06 \pm 0.03$ & $0.02 \pm 0.03$ & $0.09 \pm 0.04$ & $\mathbf{0.01 \pm 0.01}$ & $0.06 \pm 0.04$ & $\mathbf{0.02 \pm 0.03}$ & $0.02 \pm 0.02$ & $\mathbf{0.047 \pm 0.030}$ \\
 & ViT-B~\cite{kolesnikov2021image} & $0.11 \pm 0.04$ & $0.11 \pm 0.02$ & $0.19 \pm 0.03$ & $0.08 \pm 0.03$ & $0.20 \pm 0.06$ & $0.15 \pm 0.03$ & $0.03 \pm 0.04$ & $0.18 \pm 0.07$ & $\mathbf{0.01 \pm 0.01}$ & $0.117 \pm 0.042$ \\
 & Phikon-S~\cite{filiot2023scaling} & $0.12 \pm 0.01$ & $0.09 \pm 0.02$ & $0.13 \pm 0.03$ & $0.09 \pm 0.03$ & $0.09 \pm 0.08$ & $0.07 \pm 0.04$ & $0.07 \pm 0.04$ & $0.07 \pm 0.06$ & $0.17 \pm 0.08$ & $0.100 \pm 0.049$ \\
\bottomrule
\end{tabular}
}
\end{table*}
\begin{table*}[p]
  \caption{%
  Normalised differential \gls{auroc} scores for all tasks, feature extractors, downstream models, when employing \textbf{slidewise stain normalisation}~\cite{macenko2009method} at \textbf{low magnification} ($1.14$ \acs{mpp}).
  }
  \label{tbl:extended_data_norm_diff_auroc_Macenko_slidewise}
  \resizebox{\textwidth}{!}{\begin{tabular}{ll|cccc|c|cccc|c}
\toprule
 & Target & \breasticon-Subtype & \breasticon-CDH1 & \breasticon-TP53 & \breasticon-PIK3CA & \breasticon-LN status & \colonicon-MSI & \colonicon-KRAS & \colonicon-BRAF & \colonicon-SMAD4 & Average \\
Model & Feature extractor &  &  &  &  &  &  &  &  &  &  \\
\midrule
\multirow[t]{14}{*}{AttMIL} & Swin~\cite{liu2021swin} & $0.10 \pm 0.02$ & $0.23 \pm 0.03$ & $0.28 \pm 0.04$ & $0.08 \pm 0.05$ & $0.20 \pm 0.08$ & $0.20 \pm 0.04$ & $0.15 \pm 0.02$ & $0.12 \pm 0.06$ & $0.16 \pm 0.05$ & $0.170 \pm 0.048$ \\
 & CTransPath~\cite{wang2022transformer} & $0.03 \pm 0.03$ & $0.04 \pm 0.04$ & $0.05 \pm 0.03$ & $0.03 \pm 0.02$ & $0.09 \pm 0.09$ & $0.09 \pm 0.01$ & $0.08 \pm 0.04$ & $0.09 \pm 0.06$ & $0.08 \pm 0.03$ & $0.065 \pm 0.045$ \\
 & ViT-S~\cite{kolesnikov2021image} & $0.15 \pm 0.04$ & $0.13 \pm 0.04$ & $0.12 \pm 0.04$ & $0.04 \pm 0.03$ & $0.20 \pm 0.09$ & $0.20 \pm 0.05$ & $0.09 \pm 0.04$ & $0.16 \pm 0.04$ & $0.10 \pm 0.11$ & $0.133 \pm 0.061$ \\
 & Lunit-DINO~\cite{kang2023benchmarking} & $0.07 \pm 0.03$ & $0.04 \pm 0.02$ & $0.05 \pm 0.03$ & $\mathbf{0.02 \pm 0.02}$ & $0.09 \pm 0.08$ & $\mathbf{0.01 \pm 0.02}$ & $0.05 \pm 0.04$ & $\mathbf{0.02 \pm 0.03}$ & $0.09 \pm 0.06$ & $0.050 \pm 0.042$ \\
 & ViT-B~\cite{kolesnikov2021image} & $0.11 \pm 0.02$ & $0.16 \pm 0.03$ & $0.14 \pm 0.03$ & $0.04 \pm 0.03$ & $0.18 \pm 0.07$ & $0.15 \pm 0.03$ & $0.10 \pm 0.04$ & $0.11 \pm 0.04$ & $\mathbf{0.01 \pm 0.03}$ & $0.112 \pm 0.039$ \\
 & Phikon-S~\cite{filiot2023scaling} & $0.12 \pm 0.03$ & $0.09 \pm 0.04$ & $0.09 \pm 0.04$ & $0.10 \pm 0.03$ & $0.12 \pm 0.08$ & $0.07 \pm 0.03$ & $0.13 \pm 0.04$ & $0.13 \pm 0.04$ & $0.07 \pm 0.05$ & $0.102 \pm 0.043$ \\
\bottomrule
\end{tabular}
}
\end{table*}
\begin{table*}[p]
  \caption{%
  Normalised differential \gls{auroc} scores for all tasks, feature extractors, downstream models, when employing \textbf{patchwise stain normalisation}~\cite{macenko2009method} at \textbf{low magnification} ($1.14$ \acs{mpp}).
  }
  \label{tbl:extended_data_norm_diff_auroc_Macenko_patchwise}
  \resizebox{\textwidth}{!}{\begin{tabular}{ll|cccc|c|cccc|c}
\toprule
 & Target & \breasticon-Subtype & \breasticon-CDH1 & \breasticon-TP53 & \breasticon-PIK3CA & \breasticon-LN status & \colonicon-MSI & \colonicon-KRAS & \colonicon-BRAF & \colonicon-SMAD4 & Average \\
Model & Feature extractor &  &  &  &  &  &  &  &  &  &  \\
\midrule
\multirow[t]{14}{*}{AttMIL} & Swin~\cite{liu2021swin} & $0.10 \pm 0.02$ & $0.20 \pm 0.04$ & $0.26 \pm 0.03$ & $0.05 \pm 0.03$ & $0.18 \pm 0.07$ & $0.15 \pm 0.02$ & $0.13 \pm 0.04$ & $0.13 \pm 0.07$ & $0.20 \pm 0.03$ & $0.157 \pm 0.044$ \\
 & CTransPath~\cite{wang2022transformer} & $0.02 \pm 0.02$ & $0.03 \pm 0.03$ & $0.05 \pm 0.03$ & $0.04 \pm 0.01$ & $0.05 \pm 0.06$ & $0.08 \pm 0.06$ & $0.08 \pm 0.03$ & $0.08 \pm 0.03$ & $0.07 \pm 0.02$ & $0.057 \pm 0.036$ \\
 & ViT-S~\cite{kolesnikov2021image} & $0.12 \pm 0.03$ & $0.11 \pm 0.03$ & $0.13 \pm 0.04$ & $0.02 \pm 0.02$ & $0.21 \pm 0.07$ & $0.17 \pm 0.04$ & $0.08 \pm 0.06$ & $0.20 \pm 0.07$ & $0.07 \pm 0.03$ & $0.121 \pm 0.046$ \\
 & Lunit-DINO~\cite{kang2023benchmarking} & $0.06 \pm 0.03$ & $0.02 \pm 0.02$ & $0.06 \pm 0.03$ & $0.02 \pm 0.02$ & $0.08 \pm 0.06$ & $\mathbf{0.02 \pm 0.02}$ & $0.09 \pm 0.04$ & $0.05 \pm 0.06$ & $0.05 \pm 0.03$ & $0.051 \pm 0.038$ \\
 & ViT-B~\cite{kolesnikov2021image} & $0.12 \pm 0.03$ & $0.12 \pm 0.03$ & $0.14 \pm 0.02$ & $0.05 \pm 0.04$ & $0.18 \pm 0.12$ & $0.15 \pm 0.04$ & $0.11 \pm 0.04$ & $0.15 \pm 0.06$ & $\mathbf{0.02 \pm 0.02}$ & $0.118 \pm 0.054$ \\
 & Phikon-S~\cite{filiot2023scaling} & $0.13 \pm 0.02$ & $0.05 \pm 0.02$ & $0.11 \pm 0.04$ & $0.06 \pm 0.02$ & $0.11 \pm 0.09$ & $0.04 \pm 0.03$ & $0.07 \pm 0.05$ & $0.12 \pm 0.04$ & $0.07 \pm 0.06$ & $0.084 \pm 0.047$ \\
\bottomrule
\end{tabular}
}
\end{table*}
\begin{table*}[p]
  \caption{%
  Normalised differential \gls{auroc} scores for all tasks, feature extractors, downstream models, when employing \textbf{rotation/flipping augmentations} at \textbf{low magnification} ($1.14$ \acs{mpp}).
  }
  \label{tbl:extended_data_norm_diff_auroc_simple_rotate}
  \resizebox{\textwidth}{!}{\begin{tabular}{ll|cccc|c|cccc|c}
\toprule
 & Target & \breasticon-Subtype & \breasticon-CDH1 & \breasticon-TP53 & \breasticon-PIK3CA & \breasticon-LN status & \colonicon-MSI & \colonicon-KRAS & \colonicon-BRAF & \colonicon-SMAD4 & Average \\
Model & Feature extractor &  &  &  &  &  &  &  &  &  &  \\
\midrule
\multirow[t]{14}{*}{AttMIL} & Swin~\cite{liu2021swin} & $0.09 \pm 0.02$ & $0.16 \pm 0.05$ & $0.30 \pm 0.03$ & $0.07 \pm 0.02$ & $0.17 \pm 0.07$ & $0.14 \pm 0.04$ & $0.13 \pm 0.03$ & $0.10 \pm 0.04$ & $0.20 \pm 0.03$ & $0.150 \pm 0.041$ \\
 & CTransPath~\cite{wang2022transformer} & $0.04 \pm 0.03$ & $0.02 \pm 0.03$ & $0.05 \pm 0.02$ & $\mathbf{0.01 \pm 0.01}$ & $0.07 \pm 0.06$ & $0.06 \pm 0.04$ & $0.07 \pm 0.05$ & $0.08 \pm 0.02$ & $0.06 \pm 0.03$ & $0.051 \pm 0.035$ \\
 & ViT-S~\cite{kolesnikov2021image} & $0.09 \pm 0.02$ & $0.05 \pm 0.02$ & $0.16 \pm 0.04$ & $0.07 \pm 0.04$ & $0.23 \pm 0.10$ & $0.21 \pm 0.06$ & $0.05 \pm 0.04$ & $0.19 \pm 0.05$ & $0.07 \pm 0.08$ & $0.124 \pm 0.054$ \\
 & Lunit-DINO~\cite{kang2023benchmarking} & $0.09 \pm 0.02$ & $0.04 \pm 0.03$ & $0.04 \pm 0.01$ & $0.01 \pm 0.02$ & $0.08 \pm 0.07$ & $0.03 \pm 0.03$ & $0.06 \pm 0.02$ & $\mathbf{0.01 \pm 0.02}$ & $0.04 \pm 0.03$ & $0.045 \pm 0.031$ \\
 & ViT-B~\cite{kolesnikov2021image} & $0.10 \pm 0.04$ & $0.11 \pm 0.01$ & $0.17 \pm 0.02$ & $0.08 \pm 0.03$ & $0.18 \pm 0.06$ & $0.15 \pm 0.04$ & $0.09 \pm 0.08$ & $0.13 \pm 0.03$ & $\mathbf{0.02 \pm 0.03}$ & $0.114 \pm 0.042$ \\
 & Phikon-S~\cite{filiot2023scaling} & $0.10 \pm 0.02$ & $0.08 \pm 0.02$ & $0.08 \pm 0.07$ & $0.11 \pm 0.03$ & $0.09 \pm 0.07$ & $0.06 \pm 0.04$ & $0.07 \pm 0.04$ & $0.09 \pm 0.08$ & $0.19 \pm 0.09$ & $0.097 \pm 0.057$ \\
\bottomrule
\end{tabular}
}
\end{table*}
\begin{table*}[p]
  \caption{%
  Normalised differential \gls{auroc} scores for all tasks, feature extractors, downstream models, when employing \textbf{all augmentations} at \textbf{low magnification} ($1.14$ \acs{mpp}).
  }
  \label{tbl:extended_data_norm_diff_auroc_all}
  \resizebox{\textwidth}{!}{\begin{tabular}{ll|cccc|c|cccc|c}
\toprule
 & Target & \breasticon-Subtype & \breasticon-CDH1 & \breasticon-TP53 & \breasticon-PIK3CA & \breasticon-LN status & \colonicon-MSI & \colonicon-KRAS & \colonicon-BRAF & \colonicon-SMAD4 & Average \\
Model & Feature extractor &  &  &  &  &  &  &  &  &  &  \\
\midrule
\multirow[t]{14}{*}{AttMIL} & Swin~\cite{liu2021swin} & $0.07 \pm 0.01$ & $0.15 \pm 0.02$ & $0.22 \pm 0.02$ & $0.07 \pm 0.04$ & $0.15 \pm 0.08$ & $0.16 \pm 0.04$ & $0.11 \pm 0.05$ & $0.16 \pm 0.07$ & $0.17 \pm 0.05$ & $0.142 \pm 0.048$ \\
 & CTransPath~\cite{wang2022transformer} & $0.03 \pm 0.02$ & $\mathbf{0.02 \pm 0.02}$ & $\mathbf{0.01 \pm 0.01}$ & $0.04 \pm 0.03$ & $0.05 \pm 0.04$ & $0.11 \pm 0.03$ & $0.06 \pm 0.03$ & $0.10 \pm 0.06$ & $0.07 \pm 0.03$ & $0.054 \pm 0.034$ \\
 & ViT-S~\cite{kolesnikov2021image} & $0.11 \pm 0.02$ & $0.07 \pm 0.02$ & $0.14 \pm 0.02$ & $0.08 \pm 0.04$ & $0.20 \pm 0.09$ & $0.15 \pm 0.02$ & $0.05 \pm 0.03$ & $0.19 \pm 0.05$ & $0.06 \pm 0.05$ & $0.118 \pm 0.044$ \\
 & Lunit-DINO~\cite{kang2023benchmarking} & $0.08 \pm 0.03$ & $0.04 \pm 0.03$ & $0.04 \pm 0.03$ & $0.04 \pm 0.04$ & $0.08 \pm 0.06$ & $\mathbf{0.01 \pm 0.02}$ & $0.07 \pm 0.03$ & $\mathbf{0.01 \pm 0.03}$ & $0.05 \pm 0.05$ & $0.047 \pm 0.036$ \\
 & ViT-B~\cite{kolesnikov2021image} & $0.07 \pm 0.01$ & $0.11 \pm 0.04$ & $0.12 \pm 0.03$ & $0.08 \pm 0.04$ & $0.16 \pm 0.05$ & $0.14 \pm 0.04$ & $0.06 \pm 0.03$ & $0.17 \pm 0.05$ & $\mathbf{0.02 \pm 0.02}$ & $0.104 \pm 0.036$ \\
 & Phikon-S~\cite{filiot2023scaling} & $0.16 \pm 0.05$ & $0.10 \pm 0.03$ & $0.11 \pm 0.05$ & $0.12 \pm 0.05$ & $0.10 \pm 0.07$ & $0.06 \pm 0.03$ & $0.10 \pm 0.06$ & $0.18 \pm 0.07$ & $0.13 \pm 0.06$ & $0.118 \pm 0.053$ \\
\bottomrule
\end{tabular}
}
\end{table*}
\begin{table*}[p]
  \caption{%
      Test \gls{auroc} scores (averaged across the five seeds) for all tasks, feature extractors, and downstream models, when employing \textbf{no augmentations} at \textbf{low magnification} ($1.14$ \acs{mpp}).
  }
  \resizebox{\textwidth}{!}{\begin{tabular}{ll|cccc|c|cccc}
\toprule
 & Target & \breasticon-Subtype & \breasticon-CDH1 & \breasticon-TP53 & \breasticon-PIK3CA & \breasticon-LN status & \colonicon-MSI & \colonicon-KRAS & \colonicon-BRAF & \colonicon-SMAD4 \\
Model & Feature extractor &  &  &  &  &  &  &  &  &  \\
\midrule
\multirow[t]{14}{*}{AttMIL} & Swin~\cite{liu2021swin} & $0.75 \pm 0.01$ & $0.65 \pm 0.02$ & $0.54 \pm 0.02$ & $0.60 \pm 0.02$ & $0.74 \pm 0.09$ & $0.72 \pm 0.04$ & $0.51 \pm 0.05$ & $0.63 \pm 0.07$ & $0.55 \pm 0.05$ \\
 & CTransPath~\cite{wang2022transformer} & $0.82 \pm 0.02$ & $\mathbf{0.81 \pm 0.02}$ & $0.80 \pm 0.02$ & $0.62 \pm 0.01$ & $0.86 \pm 0.08$ & $0.82 \pm 0.03$ & $0.60 \pm 0.03$ & $0.71 \pm 0.01$ & $0.65 \pm 0.02$ \\
 & ViT-S~\cite{kolesnikov2021image} & $0.69 \pm 0.02$ & $0.73 \pm 0.02$ & $0.68 \pm 0.06$ & $0.58 \pm 0.04$ & $0.73 \pm 0.10$ & $0.72 \pm 0.04$ & $0.59 \pm 0.03$ & $0.58 \pm 0.03$ & $0.63 \pm 0.08$ \\
 & Lunit-DINO~\cite{kang2023benchmarking} & $0.74 \pm 0.02$ & $0.78 \pm 0.04$ & $0.79 \pm 0.03$ & $0.64 \pm 0.02$ & $0.85 \pm 0.03$ & $\mathbf{0.90 \pm 0.02}$ & $0.59 \pm 0.04$ & $\mathbf{0.76 \pm 0.04}$ & $0.69 \pm 0.02$ \\
 & ViT-B~\cite{kolesnikov2021image} & $0.74 \pm 0.04$ & $0.70 \pm 0.01$ & $0.66 \pm 0.03$ & $0.59 \pm 0.01$ & $0.74 \pm 0.06$ & $0.75 \pm 0.03$ & $0.62 \pm 0.05$ & $0.59 \pm 0.08$ & $\mathbf{0.70 \pm 0.03}$ \\
 & Phikon-S~\cite{filiot2023scaling} & $0.73 \pm 0.01$ & $0.73 \pm 0.02$ & $0.72 \pm 0.03$ & $0.57 \pm 0.02$ & $0.85 \pm 0.08$ & $0.84 \pm 0.05$ & $0.59 \pm 0.05$ & $0.70 \pm 0.06$ & $0.54 \pm 0.08$ \\
\bottomrule
\end{tabular}
}
  \label{tbl:extended_data_mean_auroc_none}
\end{table*}
\begin{table*}[p]
  \caption{%
      Test \gls{auroc} scores (averaged across the five seeds) for all tasks, feature extractors, and downstream models, when employing \textbf{slidewise stain normalisation}~\cite{macenko2009method} at \textbf{low magnification} ($1.14$ \acs{mpp}).
  }
  \label{tbl:extended_data_mean_auroc_Macenko_slidewise}
  \resizebox{\textwidth}{!}{\begin{tabular}{ll|cccc|c|cccc}
\toprule
 & Target & \breasticon-Subtype & \breasticon-CDH1 & \breasticon-TP53 & \breasticon-PIK3CA & \breasticon-LN status & \colonicon-MSI & \colonicon-KRAS & \colonicon-BRAF & \colonicon-SMAD4 \\
Model & Feature extractor &  &  &  &  &  &  &  &  &  \\
\midrule
\multirow[t]{14}{*}{AttMIL} & Swin~\cite{liu2021swin} & $0.74 \pm 0.02$ & $0.58 \pm 0.03$ & $0.55 \pm 0.03$ & $0.58 \pm 0.05$ & $0.73 \pm 0.09$ & $0.72 \pm 0.04$ & $0.55 \pm 0.01$ & $0.66 \pm 0.05$ & $0.55 \pm 0.05$ \\
 & CTransPath~\cite{wang2022transformer} & $0.82 \pm 0.04$ & $0.77 \pm 0.05$ & $0.78 \pm 0.03$ & $0.62 \pm 0.01$ & $0.84 \pm 0.10$ & $0.84 \pm 0.00$ & $0.61 \pm 0.04$ & $0.69 \pm 0.05$ & $0.63 \pm 0.03$ \\
 & ViT-S~\cite{kolesnikov2021image} & $0.70 \pm 0.04$ & $0.68 \pm 0.04$ & $0.70 \pm 0.04$ & $0.61 \pm 0.03$ & $0.74 \pm 0.10$ & $0.72 \pm 0.06$ & $0.61 \pm 0.04$ & $0.62 \pm 0.03$ & $0.62 \pm 0.13$ \\
 & Lunit-DINO~\cite{kang2023benchmarking} & $0.78 \pm 0.02$ & $0.77 \pm 0.02$ & $0.78 \pm 0.03$ & $\mathbf{0.63 \pm 0.01}$ & $0.84 \pm 0.08$ & $\mathbf{0.91 \pm 0.04}$ & $0.65 \pm 0.04$ & $\mathbf{0.76 \pm 0.06}$ & $0.62 \pm 0.07$ \\
 & ViT-B~\cite{kolesnikov2021image} & $0.74 \pm 0.02$ & $0.65 \pm 0.03$ & $0.69 \pm 0.02$ & $0.61 \pm 0.04$ & $0.75 \pm 0.08$ & $0.77 \pm 0.03$ & $0.60 \pm 0.03$ & $0.67 \pm 0.02$ & $\mathbf{0.70 \pm 0.04}$ \\
 & Phikon-S~\cite{filiot2023scaling} & $0.73 \pm 0.02$ & $0.72 \pm 0.04$ & $0.73 \pm 0.04$ & $0.55 \pm 0.03$ & $0.82 \pm 0.08$ & $0.85 \pm 0.03$ & $0.57 \pm 0.04$ & $0.65 \pm 0.01$ & $0.65 \pm 0.05$ \\
\bottomrule
\end{tabular}
}
\end{table*}
\begin{table*}[p]
  \caption{%
      Test \gls{auroc} scores (averaged across the five seeds) for all tasks, feature extractors, and downstream models, when employing \textbf{patchwise stain normalisation}~\cite{macenko2009method} at \textbf{low magnification} ($1.14$ \acs{mpp}).
  }
  \label{tbl:extended_data_mean_auroc_Macenko_patchwise}
  \resizebox{\textwidth}{!}{\begin{tabular}{ll|cccc|c|cccc}
\toprule
 & Target & \breasticon-Subtype & \breasticon-CDH1 & \breasticon-TP53 & \breasticon-PIK3CA & \breasticon-LN status & \colonicon-MSI & \colonicon-KRAS & \colonicon-BRAF & \colonicon-SMAD4 \\
Model & Feature extractor &  &  &  &  &  &  &  &  &  \\
\midrule
\multirow[t]{14}{*}{AttMIL} & Swin~\cite{liu2021swin} & $0.73 \pm 0.02$ & $0.61 \pm 0.05$ & $0.57 \pm 0.03$ & $0.60 \pm 0.03$ & $0.75 \pm 0.08$ & $0.76 \pm 0.02$ & $0.57 \pm 0.04$ & $0.65 \pm 0.08$ & $0.51 \pm 0.02$ \\
 & CTransPath~\cite{wang2022transformer} & $0.81 \pm 0.03$ & $0.78 \pm 0.04$ & $0.78 \pm 0.02$ & $0.60 \pm 0.01$ & $0.88 \pm 0.07$ & $0.83 \pm 0.06$ & $0.61 \pm 0.03$ & $0.70 \pm 0.02$ & $0.65 \pm 0.02$ \\
 & ViT-S~\cite{kolesnikov2021image} & $0.72 \pm 0.02$ & $0.70 \pm 0.03$ & $0.71 \pm 0.04$ & $0.63 \pm 0.02$ & $0.72 \pm 0.07$ & $0.75 \pm 0.04$ & $0.62 \pm 0.06$ & $0.58 \pm 0.07$ & $0.64 \pm 0.03$ \\
 & Lunit-DINO~\cite{kang2023benchmarking} & $0.77 \pm 0.02$ & $0.79 \pm 0.01$ & $0.77 \pm 0.03$ & $0.62 \pm 0.02$ & $0.85 \pm 0.07$ & $\mathbf{0.89 \pm 0.03}$ & $0.61 \pm 0.04$ & $0.73 \pm 0.07$ & $0.66 \pm 0.03$ \\
 & ViT-B~\cite{kolesnikov2021image} & $0.71 \pm 0.03$ & $0.69 \pm 0.03$ & $0.69 \pm 0.01$ & $0.60 \pm 0.05$ & $0.75 \pm 0.13$ & $0.76 \pm 0.04$ & $0.58 \pm 0.04$ & $0.63 \pm 0.06$ & $\mathbf{0.69 \pm 0.02}$ \\
 & Phikon-S~\cite{filiot2023scaling} & $0.70 \pm 0.02$ & $0.76 \pm 0.01$ & $0.72 \pm 0.04$ & $0.59 \pm 0.02$ & $0.82 \pm 0.10$ & $0.87 \pm 0.03$ & $0.62 \pm 0.05$ & $0.66 \pm 0.03$ & $0.65 \pm 0.06$ \\
\bottomrule
\end{tabular}
}
\end{table*}
\begin{table*}[p]
  \caption{%
      Test \gls{auroc} scores (averaged across the five seeds) for all tasks, feature extractors, and downstream models, when employing \textbf{rotation/flipping augmentations} at \textbf{low magnification} ($1.14$ \acs{mpp}).
  }
  \label{tbl:extended_data_mean_auroc_simple_rotate}
  \resizebox{\textwidth}{!}{\begin{tabular}{ll|cccc|c|cccc}
\toprule
 & Target & \breasticon-Subtype & \breasticon-CDH1 & \breasticon-TP53 & \breasticon-PIK3CA & \breasticon-LN status & \colonicon-MSI & \colonicon-KRAS & \colonicon-BRAF & \colonicon-SMAD4 \\
Model & Feature extractor &  &  &  &  &  &  &  &  &  \\
\midrule
\multirow[t]{14}{*}{AttMIL} & Swin~\cite{liu2021swin} & $0.76 \pm 0.02$ & $0.66 \pm 0.06$ & $0.54 \pm 0.03$ & $0.59 \pm 0.01$ & $0.77 \pm 0.08$ & $0.77 \pm 0.03$ & $0.53 \pm 0.02$ & $0.68 \pm 0.04$ & $0.52 \pm 0.03$ \\
 & CTransPath~\cite{wang2022transformer} & $0.81 \pm 0.04$ & $0.79 \pm 0.03$ & $0.80 \pm 0.01$ & $\mathbf{0.65 \pm 0.03}$ & $0.86 \pm 0.06$ & $0.85 \pm 0.03$ & $0.59 \pm 0.05$ & $0.71 \pm 0.01$ & $0.65 \pm 0.02$ \\
 & ViT-S~\cite{kolesnikov2021image} & $0.75 \pm 0.01$ & $0.77 \pm 0.02$ & $0.69 \pm 0.04$ & $0.59 \pm 0.03$ & $0.71 \pm 0.11$ & $0.70 \pm 0.06$ & $0.61 \pm 0.04$ & $0.60 \pm 0.05$ & $0.64 \pm 0.09$ \\
 & Lunit-DINO~\cite{kang2023benchmarking} & $0.76 \pm 0.02$ & $0.77 \pm 0.03$ & $0.80 \pm 0.01$ & $0.64 \pm 0.01$ & $0.86 \pm 0.07$ & $0.88 \pm 0.02$ & $0.59 \pm 0.02$ & $\mathbf{0.77 \pm 0.04}$ & $0.68 \pm 0.03$ \\
 & ViT-B~\cite{kolesnikov2021image} & $0.74 \pm 0.04$ & $0.71 \pm 0.01$ & $0.67 \pm 0.02$ & $0.58 \pm 0.03$ & $0.76 \pm 0.06$ & $0.76 \pm 0.03$ & $0.57 \pm 0.09$ & $0.66 \pm 0.03$ & $\mathbf{0.70 \pm 0.04}$ \\
 & Phikon-S~\cite{filiot2023scaling} & $0.74 \pm 0.02$ & $0.74 \pm 0.02$ & $0.76 \pm 0.08$ & $0.55 \pm 0.03$ & $0.84 \pm 0.07$ & $0.85 \pm 0.03$ & $0.59 \pm 0.04$ & $0.70 \pm 0.09$ & $0.52 \pm 0.10$ \\
\bottomrule
\end{tabular}
}
\end{table*}
\begin{table*}[p]
  \caption{%
      Test \gls{auroc} scores (averaged across the five seeds) for all tasks, feature extractors, and downstream models, when employing \textbf{all augmentations} at \textbf{low magnification} ($1.14$ \acs{mpp}).
  }
  \label{tbl:extended_data_mean_auroc_all}
  \resizebox{\textwidth}{!}{\begin{tabular}{ll|cccc|c|cccc}
\toprule
 & Target & \breasticon-Subtype & \breasticon-CDH1 & \breasticon-TP53 & \breasticon-PIK3CA & \breasticon-LN status & \colonicon-MSI & \colonicon-KRAS & \colonicon-BRAF & \colonicon-SMAD4 \\
Model & Feature extractor &  &  &  &  &  &  &  &  &  \\
\midrule
\multirow[t]{14}{*}{AttMIL} & Swin~\cite{liu2021swin} & $0.77 \pm 0.01$ & $0.66 \pm 0.02$ & $0.61 \pm 0.02$ & $0.59 \pm 0.03$ & $0.79 \pm 0.09$ & $0.74 \pm 0.04$ & $0.56 \pm 0.06$ & $0.63 \pm 0.06$ & $0.54 \pm 0.04$ \\
 & CTransPath~\cite{wang2022transformer} & $0.81 \pm 0.03$ & $\mathbf{0.79 \pm 0.02}$ & $\mathbf{0.82 \pm 0.01}$ & $0.62 \pm 0.02$ & $0.89 \pm 0.05$ & $0.79 \pm 0.03$ & $0.60 \pm 0.03$ & $0.70 \pm 0.05$ & $0.65 \pm 0.02$ \\
 & ViT-S~\cite{kolesnikov2021image} & $0.73 \pm 0.02$ & $0.73 \pm 0.02$ & $0.68 \pm 0.02$ & $0.58 \pm 0.04$ & $0.74 \pm 0.10$ & $0.75 \pm 0.02$ & $0.61 \pm 0.03$ & $0.60 \pm 0.03$ & $0.65 \pm 0.06$ \\
 & Lunit-DINO~\cite{kang2023benchmarking} & $0.76 \pm 0.03$ & $0.77 \pm 0.03$ & $0.78 \pm 0.03$ & $0.62 \pm 0.03$ & $0.86 \pm 0.06$ & $\mathbf{0.89 \pm 0.03}$ & $0.59 \pm 0.03$ & $\mathbf{0.78 \pm 0.07}$ & $0.67 \pm 0.06$ \\
 & ViT-B~\cite{kolesnikov2021image} & $0.77 \pm 0.01$ & $0.70 \pm 0.05$ & $0.70 \pm 0.03$ & $0.58 \pm 0.03$ & $0.78 \pm 0.06$ & $0.76 \pm 0.04$ & $0.60 \pm 0.02$ & $0.63 \pm 0.02$ & $\mathbf{0.70 \pm 0.04}$ \\
 & Phikon-S~\cite{filiot2023scaling} & $0.68 \pm 0.05$ & $0.71 \pm 0.03$ & $0.72 \pm 0.06$ & $0.54 \pm 0.04$ & $0.84 \pm 0.07$ & $0.84 \pm 0.03$ & $0.56 \pm 0.08$ & $0.61 \pm 0.06$ & $0.59 \pm 0.07$ \\
\bottomrule
\end{tabular}
}
\end{table*}
\begin{table*}[p]
  \caption{%
  Normalised differential \gls{auroc} scores for all tasks and feature extractors, when employing \textbf{no augmentations} at \textbf{high magnification} ($0.5$ \acs{mpp}).
  }
  \label{tbl:extended_data_norm_diff_auroc_none_high}
  \resizebox{\textwidth}{!}{\begin{tabular}{ll|cccc|c|cccc|c}
\toprule
 & Target & \breasticon-Subtype & \breasticon-CDH1 & \breasticon-TP53 & \breasticon-PIK3CA & \breasticon-LN status & \colonicon-MSI & \colonicon-KRAS & \colonicon-BRAF & \colonicon-SMAD4 & Average \\
Model & Feature extractor &  &  &  &  &  &  &  &  &  &  \\
\midrule
\multirow[t]{14}{*}{AttMIL} & Swin~\cite{liu2021swin} & $0.12 \pm 0.01$ & $0.05 \pm 0.04$ & $0.20 \pm 0.05$ & $0.08 \pm 0.02$ & $0.22 \pm 0.07$ & $0.17 \pm 0.05$ & $0.12 \pm 0.05$ & $0.13 \pm 0.04$ & $0.12 \pm 0.03$ & $0.135 \pm 0.045$ \\
 & CTransPath~\cite{wang2022transformer} & $\mathbf{0.00 \pm 0.00}$ & $0.12 \pm 0.05$ & $0.02 \pm 0.02$ & $0.06 \pm 0.03$ & $0.09 \pm 0.05$ & $0.11 \pm 0.06$ & $0.11 \pm 0.04$ & $0.09 \pm 0.03$ & $0.03 \pm 0.03$ & $0.071 \pm 0.038$ \\
 & ViT-S~\cite{kolesnikov2021image} & $0.13 \pm 0.02$ & $0.04 \pm 0.02$ & $0.19 \pm 0.07$ & $0.08 \pm 0.04$ & $0.22 \pm 0.10$ & $0.14 \pm 0.03$ & $0.12 \pm 0.02$ & $0.16 \pm 0.06$ & $0.10 \pm 0.06$ & $0.132 \pm 0.055$ \\
 & Lunit-DINO~\cite{kang2023benchmarking} & $0.06 \pm 0.02$ & $0.12 \pm 0.04$ & $0.04 \pm 0.02$ & $0.06 \pm 0.04$ & $0.05 \pm 0.02$ & $\mathbf{0.01 \pm 0.01}$ & $0.06 \pm 0.03$ & $0.03 \pm 0.03$ & $\mathbf{0.02 \pm 0.03}$ & $0.048 \pm 0.028$ \\
 & ViT-B~\cite{kolesnikov2021image} & $0.13 \pm 0.02$ & $0.05 \pm 0.02$ & $0.13 \pm 0.03$ & $0.07 \pm 0.05$ & $0.19 \pm 0.08$ & $0.08 \pm 0.04$ & $0.08 \pm 0.07$ & $0.11 \pm 0.03$ & $0.06 \pm 0.04$ & $0.100 \pm 0.047$ \\
 & Phikon-S~\cite{filiot2023scaling} & $0.04 \pm 0.01$ & $0.05 \pm 0.03$ & $0.07 \pm 0.04$ & $0.07 \pm 0.05$ & $0.03 \pm 0.02$ & $0.04 \pm 0.02$ & $0.04 \pm 0.03$ & $0.08 \pm 0.05$ & $0.10 \pm 0.06$ & $0.058 \pm 0.038$ \\
\bottomrule
\end{tabular}
}
\end{table*}
\begin{table*}[p]
  \caption{%
  Normalised differential \gls{auroc} scores for all tasks and feature extractors, when employing \textbf{patchwise stain normalisation}~\cite{macenko2009method} at \textbf{high magnification} ($0.5$ \acs{mpp}).
  }
  \label{tbl:extended_data_norm_diff_auroc_Macenko_patchwise_high}
  \resizebox{\textwidth}{!}{\begin{tabular}{ll|cccc|c|cccc|c}
\toprule
 & Target & \breasticon-Subtype & \breasticon-CDH1 & \breasticon-TP53 & \breasticon-PIK3CA & \breasticon-LN status & \colonicon-MSI & \colonicon-KRAS & \colonicon-BRAF & \colonicon-SMAD4 & Average \\
Model & Feature extractor &  &  &  &  &  &  &  &  &  &  \\
\midrule
\multirow[t]{14}{*}{AttMIL} & Swin~\cite{liu2021swin} & $0.14 \pm 0.02$ & $0.11 \pm 0.02$ & $0.21 \pm 0.04$ & $0.03 \pm 0.02$ & $0.19 \pm 0.08$ & $0.09 \pm 0.04$ & $0.18 \pm 0.10$ & $0.06 \pm 0.05$ & $0.11 \pm 0.07$ & $0.125 \pm 0.056$ \\
 & CTransPath~\cite{wang2022transformer} & $0.03 \pm 0.02$ & $0.10 \pm 0.05$ & $0.03 \pm 0.02$ & $0.06 \pm 0.02$ & $0.10 \pm 0.07$ & $0.07 \pm 0.04$ & $0.06 \pm 0.05$ & $0.11 \pm 0.03$ & $\mathbf{0.03 \pm 0.02}$ & $0.068 \pm 0.039$ \\
 & ViT-S~\cite{kolesnikov2021image} & $0.21 \pm 0.03$ & $0.03 \pm 0.03$ & $0.19 \pm 0.04$ & $0.07 \pm 0.02$ & $0.25 \pm 0.09$ & $0.12 \pm 0.04$ & $0.17 \pm 0.04$ & $0.13 \pm 0.04$ & $0.19 \pm 0.03$ & $0.149 \pm 0.044$ \\
 & Lunit-DINO~\cite{kang2023benchmarking} & $0.09 \pm 0.01$ & $0.07 \pm 0.03$ & $0.06 \pm 0.03$ & $0.05 \pm 0.02$ & $0.07 \pm 0.04$ & $0.04 \pm 0.04$ & $0.11 \pm 0.06$ & $0.04 \pm 0.06$ & $0.08 \pm 0.06$ & $0.068 \pm 0.040$ \\
 & ViT-B~\cite{kolesnikov2021image} & $0.17 \pm 0.02$ & $0.05 \pm 0.03$ & $0.15 \pm 0.03$ & $0.04 \pm 0.02$ & $0.18 \pm 0.07$ & $0.09 \pm 0.03$ & $0.08 \pm 0.06$ & $0.14 \pm 0.07$ & $0.06 \pm 0.05$ & $0.108 \pm 0.046$ \\
 & Phikon-S~\cite{filiot2023scaling} & $0.10 \pm 0.03$ & $0.05 \pm 0.04$ & $0.12 \pm 0.04$ & $0.07 \pm 0.02$ & $0.05 \pm 0.05$ & $0.06 \pm 0.04$ & $\mathbf{0.03 \pm 0.04}$ & $0.08 \pm 0.04$ & $0.05 \pm 0.03$ & $0.068 \pm 0.038$ \\
\bottomrule
\end{tabular}
}
\end{table*}
\begin{table*}[p]
  \caption{%
      Test \gls{auroc} scores (averaged across the five seeds) for all tasks and feature extractors, when employing \textbf{no augmentations} at \textbf{high magnification} ($0.5$ \acs{mpp}).
  }
  \label{tbl:extended_data_mean_auroc_none_high}
  \resizebox{\textwidth}{!}{\begin{tabular}{ll|cccc|c|cccc}
\toprule
 & Target & \breasticon-Subtype & \breasticon-CDH1 & \breasticon-TP53 & \breasticon-PIK3CA & \breasticon-LN status & \colonicon-MSI & \colonicon-KRAS & \colonicon-BRAF & \colonicon-SMAD4 \\
Model & Feature extractor &  &  &  &  &  &  &  &  &  \\
\midrule
\multirow[t]{14}{*}{AttMIL} & Swin~\cite{liu2021swin} & $0.69 \pm 0.01$ & $0.78 \pm 0.06$ & $0.63 \pm 0.05$ & $0.60 \pm 0.01$ & $0.75 \pm 0.08$ & $0.74 \pm 0.05$ & $0.55 \pm 0.06$ & $0.65 \pm 0.04$ & $0.55 \pm 0.01$ \\
 & CTransPath~\cite{wang2022transformer} & $\mathbf{0.82 \pm 0.01}$ & $0.71 \pm 0.05$ & $0.80 \pm 0.03$ & $0.62 \pm 0.03$ & $0.89 \pm 0.05$ & $0.79 \pm 0.07$ & $0.56 \pm 0.04$ & $0.69 \pm 0.02$ & $0.63 \pm 0.01$ \\
 & ViT-S~\cite{kolesnikov2021image} & $0.69 \pm 0.02$ & $0.79 \pm 0.02$ & $0.64 \pm 0.07$ & $0.60 \pm 0.03$ & $0.75 \pm 0.11$ & $0.76 \pm 0.03$ & $0.55 \pm 0.02$ & $0.62 \pm 0.07$ & $0.57 \pm 0.06$ \\
 & Lunit-DINO~\cite{kang2023benchmarking} & $0.76 \pm 0.02$ & $0.70 \pm 0.04$ & $0.79 \pm 0.02$ & $0.62 \pm 0.04$ & $0.93 \pm 0.02$ & $\mathbf{0.90 \pm 0.02}$ & $0.61 \pm 0.03$ & $0.75 \pm 0.03$ & $\mathbf{0.65 \pm 0.06}$ \\
 & ViT-B~\cite{kolesnikov2021image} & $0.69 \pm 0.02$ & $0.77 \pm 0.02$ & $0.69 \pm 0.04$ & $0.61 \pm 0.05$ & $0.79 \pm 0.09$ & $0.83 \pm 0.05$ & $0.59 \pm 0.08$ & $0.67 \pm 0.02$ & $0.61 \pm 0.03$ \\
 & Phikon-S~\cite{filiot2023scaling} & $0.78 \pm 0.01$ & $0.78 \pm 0.04$ & $0.76 \pm 0.04$ & $0.61 \pm 0.05$ & $0.95 \pm 0.02$ & $0.87 \pm 0.02$ & $0.63 \pm 0.03$ & $0.70 \pm 0.06$ & $0.57 \pm 0.06$ \\
\bottomrule
\end{tabular}
}
\end{table*}
\begin{table*}[p]
  \caption{%
      Test \gls{auroc} scores (averaged across the five seeds) for all tasks and feature extractors when employing \textbf{patchwise stain normalisation}~\cite{macenko2009method} at \textbf{high magnification} ($0.5$ \acs{mpp}).
  }
  \label{tbl:extended_data_mean_auroc_Macenko_patchwise_high}
  \resizebox{\textwidth}{!}{\begin{tabular}{ll|cccc|c|cccc}
\toprule
 & Target & \breasticon-Subtype & \breasticon-CDH1 & \breasticon-TP53 & \breasticon-PIK3CA & \breasticon-LN status & \colonicon-MSI & \colonicon-KRAS & \colonicon-BRAF & \colonicon-SMAD4 \\
Model & Feature extractor &  &  &  &  &  &  &  &  &  \\
\midrule
\multirow[t]{14}{*}{AttMIL} & Swin~\cite{liu2021swin} & $0.69 \pm 0.02$ & $0.70 \pm 0.00$ & $0.60 \pm 0.04$ & $0.63 \pm 0.01$ & $0.78 \pm 0.08$ & $0.82 \pm 0.04$ & $0.52 \pm 0.11$ & $0.71 \pm 0.06$ & $0.56 \pm 0.08$ \\
 & CTransPath~\cite{wang2022transformer} & $0.80 \pm 0.02$ & $0.70 \pm 0.05$ & $0.78 \pm 0.03$ & $0.60 \pm 0.01$ & $0.87 \pm 0.07$ & $0.84 \pm 0.04$ & $0.64 \pm 0.05$ & $0.66 \pm 0.03$ & $\mathbf{0.64 \pm 0.02}$ \\
 & ViT-S~\cite{kolesnikov2021image} & $0.62 \pm 0.03$ & $0.78 \pm 0.02$ & $0.63 \pm 0.03$ & $0.59 \pm 0.02$ & $0.73 \pm 0.10$ & $0.79 \pm 0.04$ & $0.53 \pm 0.05$ & $0.64 \pm 0.04$ & $0.48 \pm 0.03$ \\
 & Lunit-DINO~\cite{kang2023benchmarking} & $0.74 \pm 0.01$ & $0.74 \pm 0.02$ & $0.75 \pm 0.02$ & $0.61 \pm 0.02$ & $0.90 \pm 0.04$ & $0.87 \pm 0.05$ & $0.59 \pm 0.06$ & $0.73 \pm 0.07$ & $0.59 \pm 0.06$ \\
 & ViT-B~\cite{kolesnikov2021image} & $0.66 \pm 0.02$ & $0.75 \pm 0.02$ & $0.66 \pm 0.02$ & $0.62 \pm 0.01$ & $0.79 \pm 0.07$ & $0.82 \pm 0.03$ & $0.61 \pm 0.06$ & $0.63 \pm 0.08$ & $0.61 \pm 0.05$ \\
 & Phikon-S~\cite{filiot2023scaling} & $0.73 \pm 0.03$ & $0.76 \pm 0.03$ & $0.69 \pm 0.04$ & $0.59 \pm 0.02$ & $0.92 \pm 0.05$ & $0.84 \pm 0.04$ & $\mathbf{0.67 \pm 0.05}$ & $0.69 \pm 0.04$ & $0.62 \pm 0.05$ \\
\bottomrule
\end{tabular}
}
\end{table*}

\clearpage
\printbibliography
\end{refsection}
}

\end{document}